%% file: main.tex
\documentclass[runningheads]{llncs}

\usepackage[mobile,year=2026]{eccv}

\usepackage{eccvabbrv}

\usepackage{graphicx}
\usepackage{booktabs}

\usepackage[accsupp]{axessibility}  

\usepackage[utf8]{inputenc} 
\usepackage[T1]{fontenc}    
\usepackage{url}            
\usepackage{booktabs}       
\usepackage{amsfonts}       
\usepackage{nicefrac}       
\usepackage{microtype}      
\usepackage{xcolor}         
\usepackage[table,xcdraw]{xcolor} 
\usepackage{enumitem}     
\usepackage{amsfonts}
\usepackage{graphicx}
\usepackage{subcaption} 
\usepackage{multirow}
\usepackage{array}   
\usepackage{adjustbox}
\definecolor{objblue}{RGB}{3,139,221}  
\definecolor{attrred}{RGB}{255,67,67}    
\definecolor{easygreen}{RGB}{0,156,75}  
\definecolor{middleyellow}{RGB}{242,89,34}  
\definecolor{hardred}{RGB}{216,56,58}   
\usepackage{makecell}
\usepackage[most]{tcolorbox}
\usepackage{pifont}   

\usepackage{hyperref}

\usepackage{orcidlink}

\begin{document}

\title{%
  \raisebox{-0.6ex}{\includegraphics[height=3.4ex]{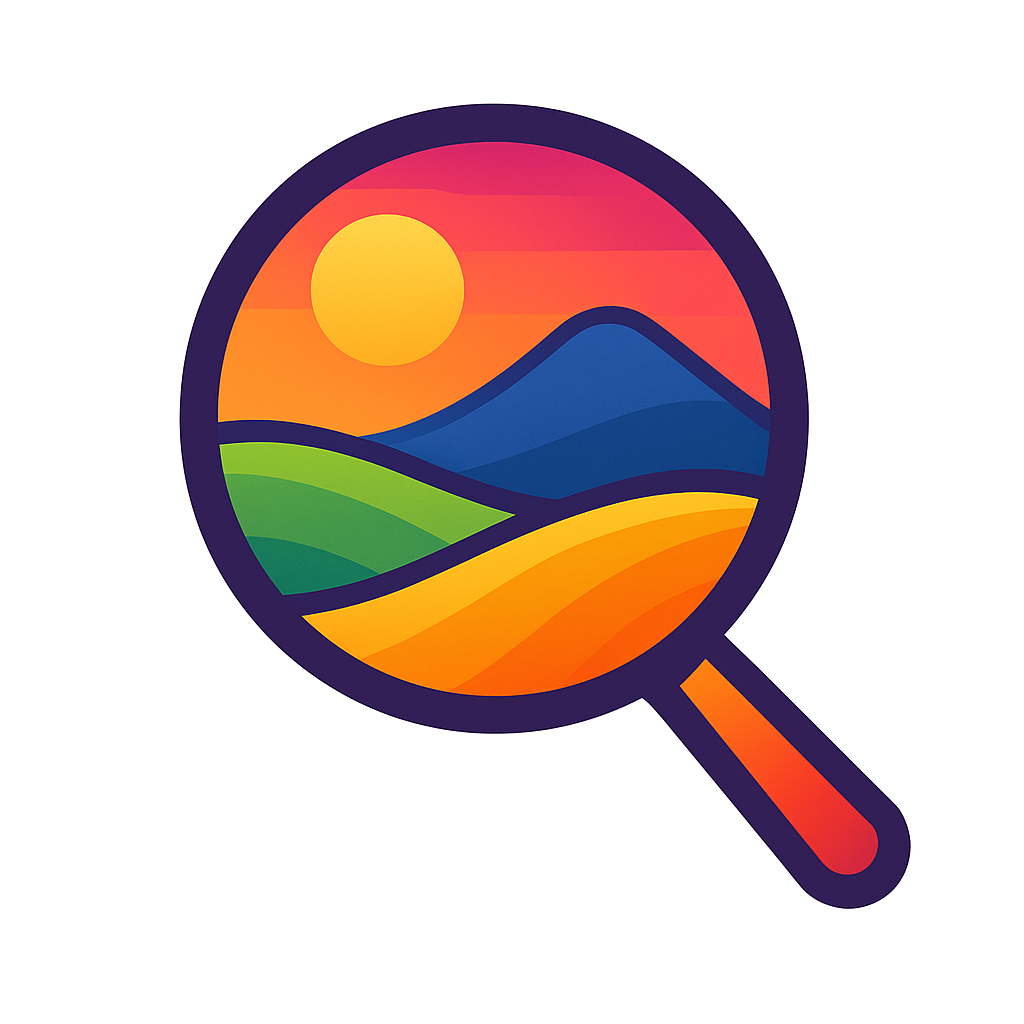}}\hspace{0.1em}%
  TIIF-Bench: How Does Your T2I Model \\ Follow Your Instructions?
}

\titlerunning{TIIF-Bench}

\author{%
{\fontsize{8}{10}\selectfont
Xinyu Wei\inst{1,3} \and
Jinrui Zhang\inst{1,3} \and
Zeqing Wang\inst{1,3} \and
Hongyang Wei\inst{2,3} \\
Zhen Guo\inst{1,3} \and
Bairui Li\inst{1,3} \and
Lei Zhang\inst{1,3}
}%
}

\institute{%
{\fontsize{7}{9}\selectfont
The Hong Kong Polytechnic University \and
Tsinghua University \and
OPPO Research Institute
}%
}

\maketitle

\begin{abstract}
The rapid advancements of Text-to-Image (T2I) models have ushered in a new phase of AI-generated content, marked by their growing ability to interpret and follow user instructions.
However, existing T2I model evaluation benchmarks fall short in limited prompt diversity and complexity, as well as coarse evaluation metrics, making it difficult to evaluate the fine-grained alignment performance between textual instructions and generated images.
In this paper, we present \textbf{TIIF-Bench} (\textbf{T}ext-to-\textbf{I}mage \textbf{I}nstruction \textbf{F}ollowing \textbf{Bench}mark), aiming to systematically assess T2I models’ ability in interpreting and following intricate textual instructions. 
TIIF-Bench comprises 5,000 prompts organized along multiple dimensions and categorized into three levels of difficulty and complexity. To rigorously evaluate robustness to prompt length, each prompt is provided in both short and long versions with identical core semantics. We further propose a novel Global Normalized Edit Distance (GNED) metric for text rendering and provide aspect-ratio-diverse reference images for each prompt to assess style control. In addition, we collect 100 high-quality designer-level prompts covering diverse scenarios for comprehensive evaluation.
To enable scalable and fine-grained evaluation, we explore the best paradigm for leveraging the world knowledge encoded in large Vision-Language Models (VLMs) as automated binary evaluators. Through extensive ablations, we develop a fully reproducible evaluator that provides interpretable reasoning and reliable verification, enabling our benchmark to discern subtle variations in T2I model outputs.
Through comprehensive benchmarking of mainstream T2I models on TIIF-Bench, we analyze the strengths and weaknesses of current T2I systems and reveal the limitations of existing evaluation benchmarks.

\end{abstract}

\begin{figure}[t]
  \centering
  \includegraphics[width=\linewidth]{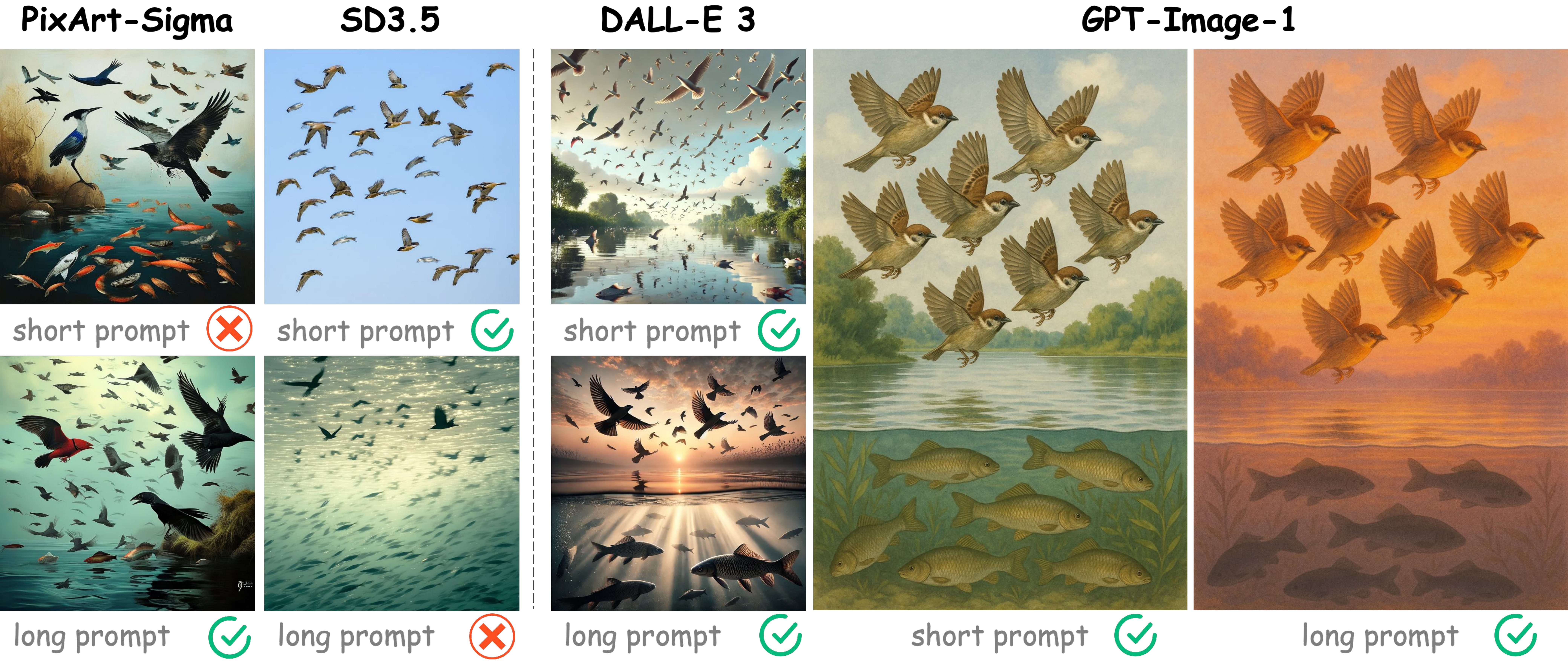}
  \caption{Prompt‑length sensitivity example.  
  Short prompt: \textcolor[rgb]{0.5,0.5,0.5}{\textit{“The birds are more numerous than the fish.”}}  
  Long prompt: \textcolor[rgb]{0.5,0.5,0.5}{\textit{“The birds, with their feathers catching the gentle light of dawn, vastly outnumber their aquatic counterparts, the fish, which glide silently beneath the rippling surface of the water, their sleek forms moving like shadows in the depths below.”}} PixArt-Sigma and SD 3.5 show clear sensitivity to prompt length, producing different results for semantically equivalent prompts. 
In contrast, DALL·E 3 and GPT-Image-1 remain robust, maintaining consistent instruction-following performance.}
  \label{fig:short-long-ablation}
  \vspace{-4mm}
\end{figure}

\section{Introduction}
Text‑to‑Image (T2I) generation has emerged as a cornerstone of multimodal AI, enabling the translation of abstract textual concepts into detailed visual content, advancing applications from digital art to scientific visualization. Recent T2I models can be categorized into 3 main paradigms. 
Diffusion-based methods—exemplified by Stable Diffusion\cite{rombachHighResolutionImageSynthesis2022, esserScalingRectifiedFlow2024}, PixArt\cite{chenPixArtSWeaktoStrongTraining2024, chenPixArt$a$FastTraining2023}, FLUX\cite{flux2024}, SANA\cite{chenSANASprintOneStepDiffusion2025, xieSANA15Efficient2025}, Qwen-Image\cite{wu2025qwenimagetechnicalreport}, and others\cite{betkerImprovingImageGeneration, imagen-team-googleImagen32024, zhuoLuminaNextMakingLuminaT2X2024, liHunyuanDiTPowerfulMultiResolution2024, Cui_2024_CVPR, tong2026delving, lin2025pixwizard}—leverage U-Net or Diffusion-Transformer backbones to iteratively denoise Gaussian noise into photorealistic images, achieving strong visual fidelity and diversity.
Autoregressive (AR) approaches, such as LlamaGen \cite{sunAutoregressiveModelBeats2024}, Infinity\cite{Infinity}, VAR\cite{VAR}, and other influential open-source models, treat images as token sequences and synthesize them through next-token prediction or scale-progressive generation.
Unified-model (UM) approaches, including the Show-o series\cite{xieShowoOneSingle2024, xie2025showo2improvednativeunified}, Janus\cite{wuJanusDecouplingVisual2024} and JanusPro\cite{chenJanusProUnifiedMultimodal}, the Lumina-mGPT series\cite{liu2025_Luminamgpt_illuminateflexiblephotorealistic, xin2025luminamgpt20standaloneautoregressive}, Bagel\cite{deng2025emergingpropertiesunifiedmultimodal} and the EMU series\cite{cui2025emu35nativemultimodalmodels}, unify generation and understanding within a single transformer backbone, aiming to mutually enhance multimodal reasoning and image synthesis.
Very recently, commercial T2I systems such as the Nano-Banana series\cite{nano-banana, nano-banana-pro, nano-banana-2}, GPT-Image-1\cite{hurst2024gpt}, Imagen 3\cite{baldridge2024imagen}, and MidJourney\cite{Midjourney} have further propelled T2I generation to a new level. In particular, GPT-Image-1\cite{hurst2024gpt} demonstrates strong instruction-following capability, enabling it to interpret highly intricate prompts and produce visually precise, stylistically coherent images within a single conversational interaction. Meanwhile, the Nano-Banana series\cite{nano-banana, nano-banana-pro, nano-banana-2} pushes the boundaries of generative reasoning and world knowledge integration in T2I models.

With the rapid development of T2I techniques, how to comprehensively evaluate the performance, especially the instruction-following capability, of modern T2I models has become an important issue. 
Existing research on the evaluation of T2I models generally falls into two complementary categories.
\textbf{Preference-alignment approaches}, such as {VQAScore}~\cite{lin2024evaluatingtexttovisualgenerationimagetotext}, {HPSv2}~\cite{wu2023humanpreferencescorev2}, HPSv3~\cite{ma2025hpsv3widespectrumhumanpreference}, {VisionReward}~\cite{xu2025visionrewardfinegrainedmultidimensionalhuman}, UnifiedReward~\cite{wang2026unifiedrewardmodelmultimodal}, and UnifiedReward-CoT~\cite{wang2025unifiedmultimodalchainofthoughtreward}, leverage learned reward models to align evaluation with human preferences.
\textbf{Benchmark-driven approaches}, exemplified by TIFA~\cite{hu2023tifaaccurateinterpretabletexttoimage}, DSG-Bench~\cite{cho2024davidsonianscenegraphimproving}, DPG-Bench~\cite{hu2024ellaequipdiffusionmodels}, Gecko~\cite{wiles2025revisitingtexttoimageevaluationgecko}, {CompBench++}~\cite{huang2025t2icompbenchenhancedcomprehensivebenchmark}, {GenEval}~\cite{ghosh2023genevalobjectfocusedframeworkevaluating}, and {GenAI Bench}~\cite{li2024genaibenchevaluatingimprovingcompositional}, construct structured prompt sets across compositional dimensions (\textit{e.g.}, object attributes, relations, and numeracy) and employ CLIP-based metrics for automatic evaluation.
With the emergence of more capable models such as GPT-Image-1\cite{hurst2024gpt}, the overall capability of T2I systems has significantly improved, leading to a new generation of more challenging and comprehensive benchmarks. Recent works—including OneIG-Bench~\cite{chang2025oneigbenchomnidimensionalnuancedevaluation}, LongT2I-Bench~\cite{yang2025longt2ibenchbenchmarkevaluatinglong}, and GenEval2~\cite{kamath2025geneval2addressingbenchmark}, among others~\cite{ye2025echo4oharnessingpowergpt4o}—introduce more complex evaluation settings, while some studies further incorporate world knowledge and reasoning into the evaluation process~\cite{chen2025r2ibenchbenchmarkingreasoningdriventexttoimage, niu2025wiseworldknowledgeinformedsemantic, sun2025t2ireasonbenchbenchmarkingreasoninginformedtexttoimage, li2025easierpaintingthinkingtexttoimage, li2026easierpaintingthinkingtexttoimage, wang2025videoverse, wang2026timecausality}.

\begin{figure}[t]
  \centering
  \includegraphics[width=0.999\linewidth]{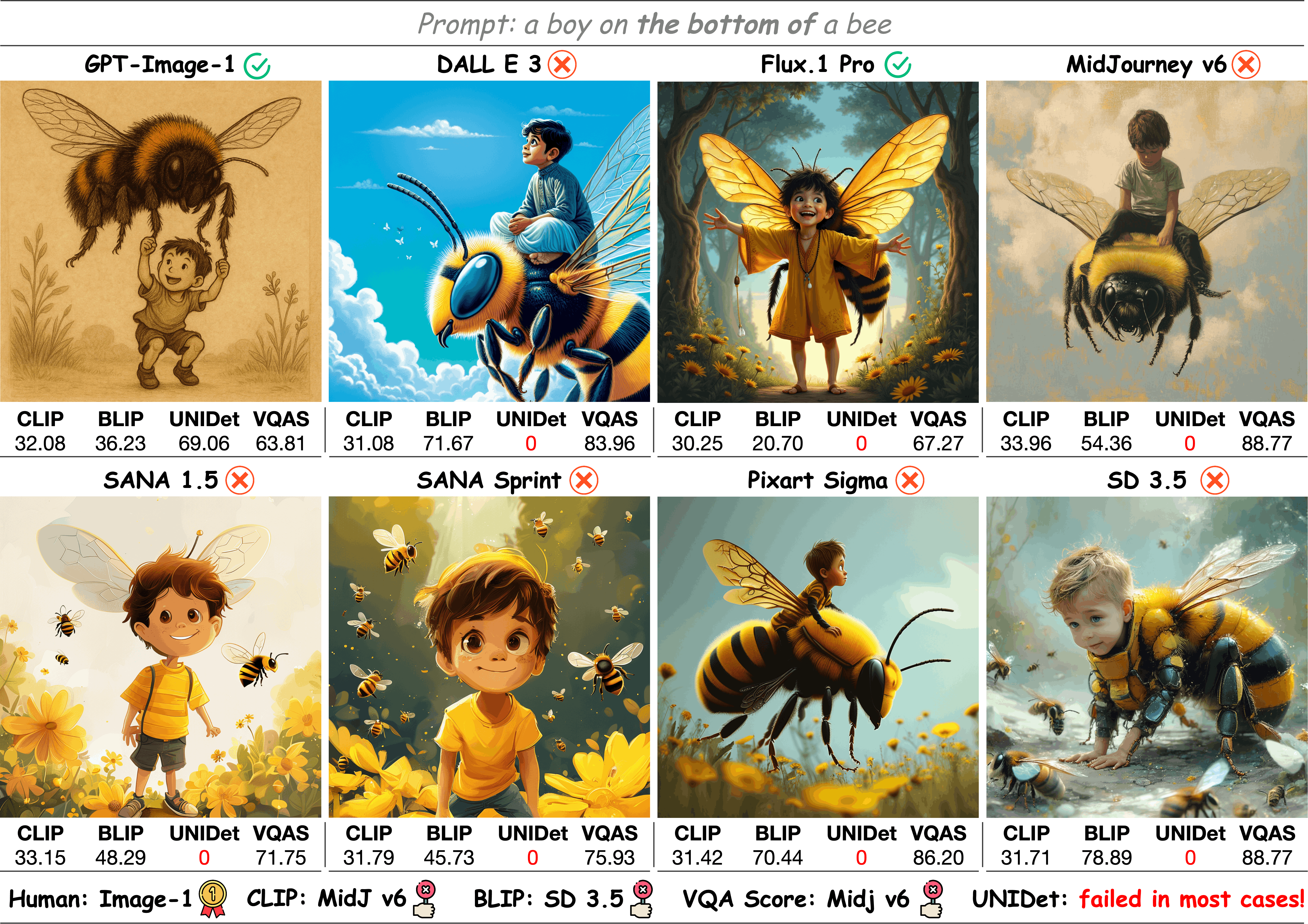}
  \caption{
Limitations of expert scorers widely used in classic T2I benchmarks. 
CLIP~\cite{radford2021learningtransferablevisualmodels}, BLIP~\cite{li2022blipbootstrappinglanguageimagepretraining}, and VQAScore~\cite{lin2024evaluatingtexttovisualgenerationimagetotext} capture only coarse image-text alignment and struggle to evaluate fine-grained instruction following. 
UniDet~\cite{zhou2022simplemultidatasetdetection} often fails to detect objects in complex generated scenes, leading to scores inconsistent with human judgments.
}
  \label{fig:other-eval-wrong}
  \vspace{-4mm}
\end{figure}

While existing benchmarks have advanced T2I evaluation, they still exhibit several limitations.
\textbf{First}, although T2I models often show different performance when given prompts with identical semantics but different lengths, most benchmarks provide only a single prompt version, preventing systematic study of prompt-length robustness (Fig.~\ref{fig:short-long-ablation}).
\textbf{Second}, many benchmarks contain substantial semantic redundancy, limiting their ability to evaluate model generalization across diverse concepts (Fig.~\ref{fig:other-bench-analyze}).
\textbf{Finally}, current evaluation protocols remain inadequate for fine-grained instruction following. Traditional expert scorers such as CLIP or BLIP capture only coarse image–text alignment and often disagree with human judgments (Fig.~\ref{fig:other-eval-wrong}). Some recent works employ large vision-language models (VLMs) as evaluators \cite{huang2025t2icompbenchenhancedcomprehensivebenchmark}, but their evaluation queries are typically too coarse to fully exploit VLM reasoning ability. In addition, several benchmarks rely on closed-source APIs \cite{wu2024conceptmixcompositionalimagegeneration, ye2025echo4oharnessingpowergpt4o, li2026easierpaintingthinkingtexttoimage}, making evaluation difficult to reproduce. Directly applying general-purpose open-source VLMs (e.g., Qwen2.5-VL \cite{qwen2025qwen25technicalreport}) also yields unreliable judgments without task-specific adaptation.

\begin{figure}[t]
  \centering
  \includegraphics[width=\linewidth]{figs/semantic_uni_bar.pdf}
  \vspace{-6mm}
  \caption{
Semantic uniqueness after de-duplication with a cosine similarity threshold of 0.85. 
Many existing benchmarks contain substantial semantic redundancy, with fewer than 60\% semantically unique prompts. 
While Gecko\cite{wiles2025revisitingtexttoimageevaluationgecko} and ConceptMix\cite{wu2024conceptmixcompositionalimagegeneration} achieve higher semantic uniqueness, they mainly focus on compositional evaluation. 
In contrast, TIIF-Bench achieves the highest semantic uniqueness while covering multiple dimensions, including composition, reasoning, text rendering, and style control.
}
  \label{fig:other-bench-analyze}
  \vspace{-4mm}
\end{figure}

To address these challenges, we introduce TIIF-Bench (\textbf{T}ext-To-\textbf{I}mage \textbf{I}nstruction \textbf{F}ollowing \textbf{Bench}mark), a benchmark designed for fine-grained evaluation of instruction-following capability in T2I models. We construct prompts using an attribute-composition strategy, combining ten concept pools with six compositional dimensions to ensure semantic uniqueness and diverse sentence structures. Two additional dimensions—text rendering and style control—are introduced to evaluate capabilities beyond traditional compositional reasoning. The benchmark also includes 100 designer-level prompts capturing real-world human aesthetic priors. Each prompt is provided in both short and extended versions to evaluate robustness to prompt length, resulting in a total of 5,000 prompts.
In our evaluation framework, each prompt is paired with a set of attribute-specific yes/no questions that serve as a checklist, enabling vision-language models (VLMs) to perform fine-grained verification rather than producing a single coarse score. 
We further investigate best paradigms for employing VLMs as automated evaluators and develop our own TIIF-Evaluator (see Section~\ref{sec:Ablation and Training of the VLM Evaluator}).
In addition, text rendering accuracy is measured using the proposed Global Normalized Edit Distance (GNED) metric, while style control is evaluated through similarity with reference images using a weighted combination of CSD \cite{somepalli2024measuringstylesimilaritydiffusion}, DINOv3 \cite{dinov3}, and SigLIP2 \cite{tschannen2025siglip2multilingualvisionlanguage}.

We benchmark 
popular closed-source models, including Nano-Banan series\cite{nano-banana, nano-banana-pro}, GPT-Image-1 \cite{hurst2024gpt},  DALLE-3 \cite{betker2023improving}, MidJourney V7 \cite{Midjourney}, 
alongside leading open-source diffusion-based models\cite{wu2025qwenimagetechnicalreport, qin2025luminaimage20unifiedefficient, rombachHighResolutionImageSynthesis2022, esserScalingRectifiedFlow2024, chenPIXARTdFastControllable2024, chenPixArtSWeaktoStrongTraining2024, xieSANA15Efficient2025, chenSANASprintOneStepDiffusion2025, flux2024,flux.2},
and UM pipelines\cite{xie2025showo2improvednativeunified, chenJanusProUnifiedMultimodal, liu2025_Luminamgpt_illuminateflexiblephotorealistic, xin2025luminamgpt20standaloneautoregressive, deng2025emergingpropertiesunifiedmultimodal, cui2025emu35nativemultimodalmodels}. 
Our analysis reveals several notable findings. Models that achieve higher overall scores on TIIF-Bench tend to exhibit stronger robustness to prompt length variations, whereas lower-performing models are significantly more sensitive to such changes. This observation suggests a positive correlation between a model’s ability to comprehend textual instructions and the quality of its generated images.
Furthermore, although UM approaches generally produce images with lower visual fidelity compared to sota diffusion models, their instruction-following performance is often comparable. This result highlights a key advantage of unified architectures, where integrating understanding and generation within a single transformer backbone allows multimodal reasoning to directly support image synthesis.
The contributions of this paper are summarized as follows:

\textbf{(i) Prompt-design methodology.}
We identify critical limitations in existing T2I benchmarks, including fixed-length prompts, high semantic redundancy, and limited syntactic diversity. To address these issues, we propose a novel attribute-composition strategy that improves semantic coverage and diversity, and introduce additional evaluation dimensions together with prompt-length variations to systematically assess model robustness.

\textbf{(ii) Fine-grained evaluation protocol.}
We investigate effective paradigms for using VLMs as automated evaluators and develop a dedicated TIIF-Evaluator. Leveraging the world knowledge of VLMs, the evaluator performs attribute-specific yes/no verification with interpretable chain-of-thought reasoning, enabling more fine-grained and reliable alignment assessment than previous metrics.

\textbf{(iii) Empirical insights for T2I evaluation.}
Through extensive experiments on TIIF-Bench, we uncover consistent patterns in how T2I models with different architectures follow instructions during image generation, providing practical insights for future research on instruction-following image synthesis.

\section{Related Work}
\vspace{-2mm}
T2I models have achieved remarkable progress in generating high-quality images. However, how to evaluate their performance, particularly in terms of the instruction-following capability, remains a challenging task. Existing work in T2I evaluation generally falls into two streams: (i) scoring models aligned with human visual preferences, and (ii) benchmarks for structured evaluation.

\textbf{Scoring Methods Aligned with Human Preferences.}
CLIPScore~\cite{hesselCLIPScoreReferencefreeEvaluation2022} is a widely adopted metric for evaluating image–text alignment, but it often fails on complex prompts due to CLIP’s inherent bag-of-words behavior. 
To address this limitation, VQAScore~\cite{lin2024evaluatingtexttovisualgenerationimagetotext} fine-tunes a CLIP-FlanT5 specifically for T2I image quality evaluation. 
HPDv2~\cite{wu2023humanpreferencescorev2} mitigates biases in human preference data. When used to fine-tune CLIP, it results in HPSv2, which demonstrates improved alignment with human judgments.
However, these approaches remain fundamentally constrained by CLIP-based representations: they struggle to capture fine-grained semantic correspondence. 
VisionReward~\cite{xu2025visionrewardfinegrainedmultidimensionalhuman} attempts to address this limitation by learning reward model through hierarchical visual assessment and interpretable weighting. 
Nevertheless, it is primarily designed for video quality evaluation, which limits its applicability to static T2I tasks. 
More recently, the UnifiedReward series~\cite{wang2026unifiedrewardmodelmultimodal, wang2025unifiedmultimodalchainofthoughtreward} employs vision-language models (VLMs) to evaluate overall image–text alignment. However, these approaches typically produce holistic scores for entire images and cannot provide fine-grained verification at the level of specific semantic attributes or binary questions.

\textbf{Benchmark-Driven Evaluation Frameworks.}
Early benchmark-driven evaluation frameworks, such as TIFA~\cite{hu2023tifaaccurateinterpretabletexttoimage}, DSG-Bench~\cite{cho2024davidsonianscenegraphimproving}, DPG-Bench~\cite{hu2024ellaequipdiffusionmodels}, Gecko~\cite{wiles2025revisitingtexttoimageevaluationgecko}, {CompBench++}~\cite{huang2025t2icompbenchenhancedcomprehensivebenchmark}, {GenEval}~\cite{ghosh2023genevalobjectfocusedframeworkevaluating}, and {GenAI Bench}~\cite{li2024genaibenchevaluatingimprovingcompositional}, construct structured prompt sets across compositional dimensions (e.g., object attributes, relations, and numeracy) and typically rely on CLIP-based metrics or relatively weak VQA models, such as mPLUG~\cite{li2022mplugeffectiveefficientvisionlanguage}, for automatic evaluation.
As T2I models have rapidly improved—particularly with the emergence of systems such as GPT-Image-1~\cite{hurst2024gpt}—recent benchmarks have become increasingly comprehensive and challenging. Works such as OneIG-Bench~\cite{chang2025oneigbenchomnidimensionalnuancedevaluation}, LongT2I-Bench~\cite{yang2025longt2ibenchbenchmarkevaluatinglong}, and GenEval2~\cite{kamath2025geneval2addressingbenchmark}, among others~\cite{ye2025echo4oharnessingpowergpt4o, an2026genius, wei2026mico}, introduce more complex evaluation settings. In addition, several studies incorporate world knowledge and reasoning into T2I evaluation, including WISE~\cite{niu2025wiseworldknowledgeinformedsemantic}, T2I-Bench~\cite{chen2025r2ibenchbenchmarkingreasoningdriventexttoimage}, and T2IReasonBench~\cite{sun2025t2ireasonbenchbenchmarkingreasoninginformedtexttoimage}, as well as other benchmarks emphasizing reasoning capabilities~\cite{li2025easierpaintingthinkingtexttoimage, li2026easierpaintingthinkingtexttoimage}.

\section{Construction of TIIF-Bench}
\subsection{Limitations of Current T2I Evaluation Benchmarks}
\label{subsec: Limitations of Current T2I Model Evaluation Benchmarks}
As T2I models continue to evolve, accurately evaluating their ability to follow natural language instructions has become increasingly important. However, existing benchmarks still suffer from several limitations, which can be broadly categorized into \textbf{prompt-design flaws} and \textbf{evaluation-protocol flaws}.

\textbf{Prompt Design Flaws}.
We identify two major issues in the prompt design of existing benchmarks.
\textbf{First}, many benchmarks rely on templated prompts with fixed or narrowly distributed lengths. For example, in \textsc{CompBench++}, the 2D and 3D evaluation dimensions contain a total of 2,000 prompts, yet their lengths are restricted to only four values (5, 6, 7, and 8 words). Such limited variation prevents evaluation of model robustness to prompt length, despite empirical evidence that T2I models can be highly sensitive to prompt length variations, as shown in Fig.~\ref{fig:short-long-ablation}.
\textbf{Second}, existing benchmarks often exhibit substantial semantic redundancy. To quantify this issue, we extract CLIP~\cite{radford2021learningtransferablevisualmodels} text embeddings for all prompts and compute pairwise cosine similarity. As shown in Fig.~\ref{fig:other-bench-analyze}, many benchmarks contain a large proportion of semantically redundant prompts.

\begin{figure}[t]
  \centering
  \includegraphics[width=\linewidth]{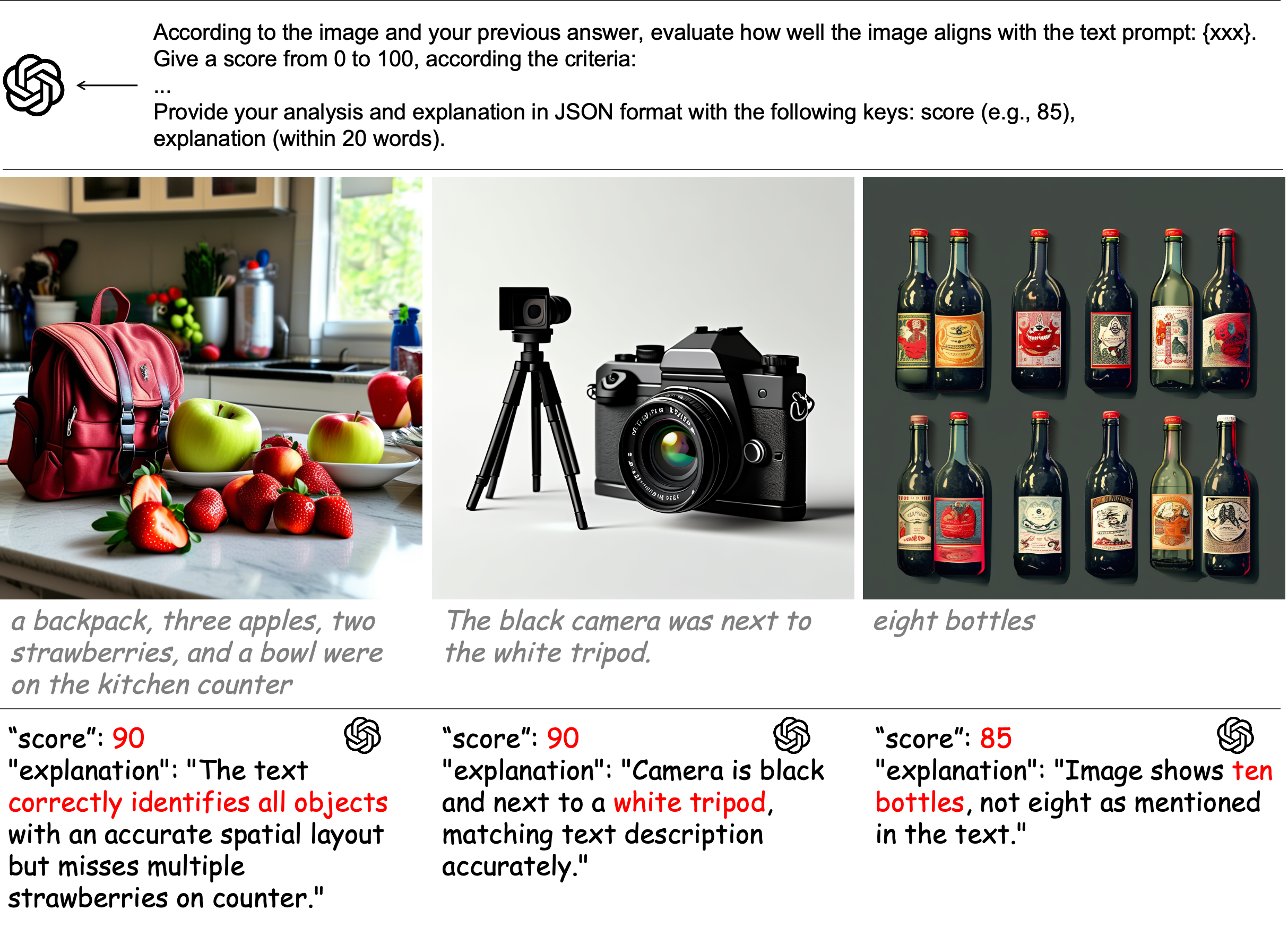}
  \vspace{-3mm}
  \caption{Failure cases of coarse prompt-based evaluation with closed-source VLMs.
  Existing methods often concatenate the full T2I prompt with a generic evaluation query.
  Because the caption is explicitly exposed to the evaluator, VLMs such as GPT-4o\cite{hurst2024gpt} may
  hallucinate unseen attributes or relations from the textual prior, leading to overly
  optimistic judgments that are not fully grounded in the image content.}
  \label{fig:gpt-eval-wrong}
\end{figure}

\textbf{Evaluation Protocol Flaws}.
We further identify two limitations in current evaluation protocols.
\textbf{First, reliance on weak expert models.}
Most benchmarks rely on CLIP~\cite{radford2021learningtransferablevisualmodels} to measure image–text alignment. However, CLIP is known to behave similarly to a “bag-of-words” model~\cite{lin2024evaluatingtexttovisualgenerationimagetotext}, often failing to distinguish semantically different prompts such as “a boy is on the bottom of a bee” and “a bee is on the bottom of a boy”~\cite{lin2025draw, lin2026perceive, zhang2025mavis}. Other expert models, such as BLIP~\cite{li2022blipbootstrappinglanguageimagepretraining}, offer limited improvements for fine-grained alignment. Although recent grounding and perception systems improve fine-grained localization and understanding~\cite{wang2026locateanything, wang2024mr}, conventional object detectors such as UniDet~\cite{zhou2022simplemultidatasetdetection} still frequently struggle with complex scenes generated by modern T2I models (see Fig.~\ref{fig:other-eval-wrong}).
\textbf{Second, coarse utilization of VLM evaluators.}
Some recent benchmarks employ strong VLMs to assess image–text alignment. However, these evaluations typically rely on a single generic query, such as \textit{“Evaluate how well the image aligns with the text prompt: \{xxx\}.”}~\cite{huang2025t2icompbenchenhancedcomprehensivebenchmark}. Such coarse queries fail to decompose the rich multi-attribute semantics present in modern T2I prompts. Moreover, explicitly including the full T2I prompt (\textit{i.e.}, the caption) in the evaluation question can induce hallucinations in VLM responses, as illustrated in Fig.~\ref{fig:gpt-eval-wrong}.

As illustrated in Fig.~\ref{fig:gpt-eval-wrong}, several recent T2I benchmarks evaluate image--text alignment by directly inserting the full T2I prompt into a generic query posed to a closed-source VLM. Although this strategy appears simple, it introduces a systematic source of bias. In particular, when the evaluation prompt explicitly exposes the target caption, the VLM may rely excessively on the textual prior rather than the visual evidence, thereby hallucinating missing attributes or relations and assigning overly optimistic scores. This issue becomes especially severe for modern T2I prompts that are long, compositional, and semantically dense, where a single coarse query is insufficient to disentangle the underlying instruction constraints.
Our observations suggest that such caption-conditioned coarse prompting is not merely low-resolution, but can also compromise evaluator reliability. This motivates the attribute-specific binary-question protocol adopted in TIIF-Bench, which avoids querying the evaluator with the entire prompt as a single holistic instruction and instead decomposes evaluation into fine-grained, verifiable subproblems.

In addition, some benchmarks rely on closed-source APIs as evaluators, making their evaluation pipelines difficult to reproduce. Others directly adopt general-purpose VLMs (e.g., Qwen2.5-VL~\cite{qwen2025qwen25technicalreport}) as off-the-shelf evaluators. However, our experiments indicate that without task-specific adaptation, such models struggle to assess the complex prompts in TIIF-Bench.

\begin{figure}[t]
  \centering
  \includegraphics[width=0.95\linewidth]{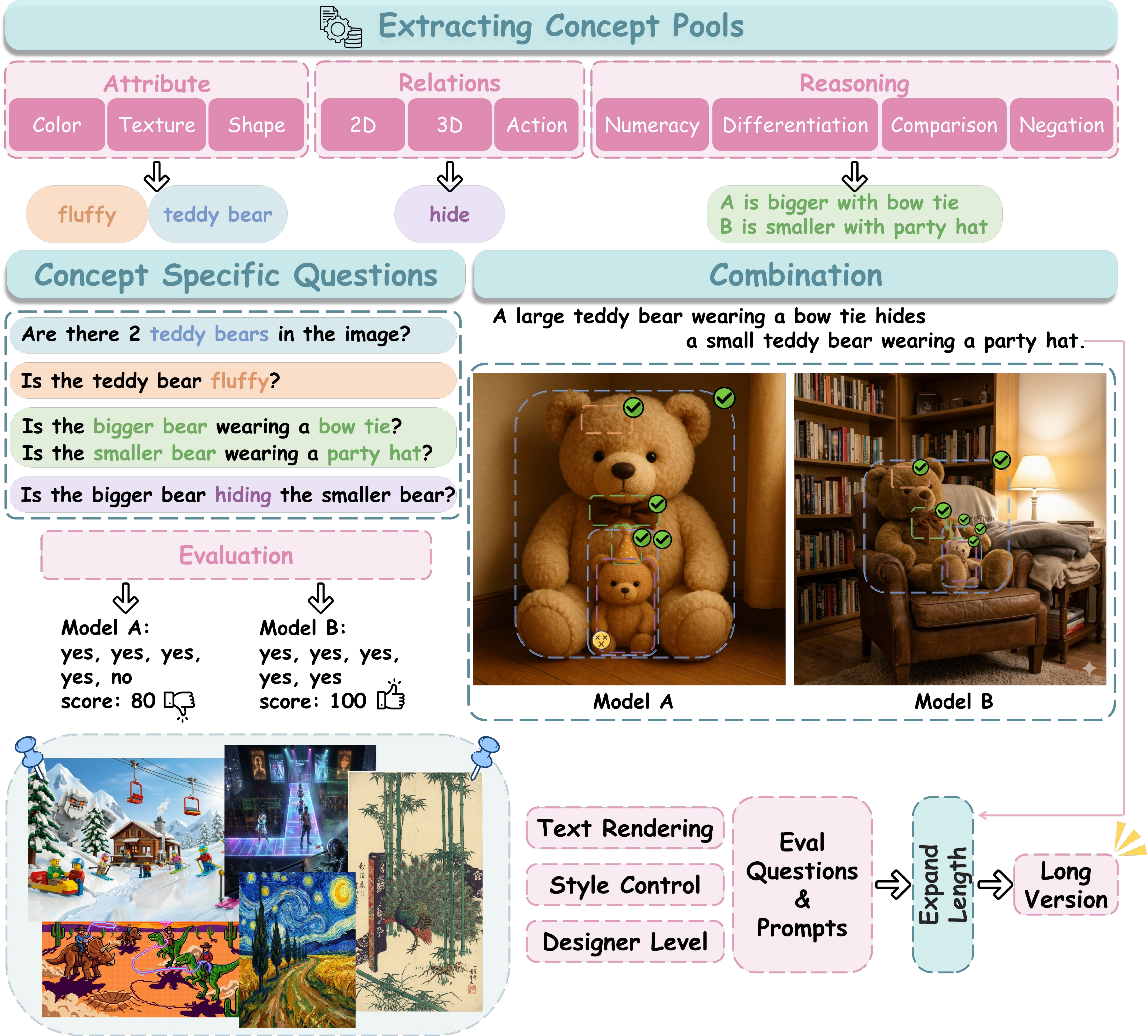}
  \caption{
Prompt construction and evaluation pipeline of TIIF-Bench. 
In addition to systematically reusing prompts from existing benchmarks, we incorporate text-rendering prompts from Lex-Art~\cite{zhao2025lexartrethinkingtextgeneration}, as well as curated style-control prompts (each paired with reference images of multiple aspect ratios) and real-world designer-level prompts. 
All prompts are further expanded to produce long-form counterparts.
}
  \label{fig: data-pipeline}
  \vspace{-5mm}
\end{figure}

\subsection{Prompts of \textsc{TIIF‑Bench}}

To address the limitations of existing T2I evaluation benchmarks, we build upon the hierarchical prompt taxonomies of prior work by introducing a two-stage process: \textbf{concept pool construction} followed by \textbf{attribute composition}. This process generates new prompts that continue to target core capabilities such as object-attribute binding and spatial layout.
To more comprehensively assess instruction-following performance, we further introduce three new evaluation dimensions: text rendering, style control, and real-world designer-level prompts.
Each prompt is also processed by a length augmentation module that generates both concise and verbose versions, enabling evaluation across varying linguistic lengths. The full data generation pipeline is illustrated in Fig.~\ref{fig: data-pipeline}.

\textbf{Concept Pool Construction}.
We first group the prompts in existing benchmarks based on their semantics and leverage GPT-4o to extract the underlying \emph{object–attribute/relation pairs}, forming a set of dimension-specific concept pools. In total, we construct \textbf{10 concept pools} from existing benchmarks, categorized them into \textbf{three groups}, as summarized in Tab. A1 in \textbf{Appendix}.

\textbf{Attribute Composition}. 
Building upon concept pools, we generate prompts by randomly combining attributes from each pool, leveraging GPT-4o to compose them into natural instructions. As illustrated in Tab.~\ref{tab:concept-compse}, we define \textbf{36 distinct combinations}, each paired with a dedicated meta-prompt to guide GPT-4o to assemble instructions. 
This sampling strategy ensures that the resulting prompts are both semantically unique and compositionally diverse.
Prompts that combine elements drawn from a \emph{single} concept-pool group are classified as \textbf{Basic Following}. In contrast, \textbf{Advanced Following} prompts intertwine elements taken from \emph{different} concept-pool groups, yielding more intricate compositions.

\begin{table}[t]
\captionsetup{skip=4pt}
\caption{Based on difficulty, we define \textbf{36 attribute-pool combinations}, grouped into three levels: \textcolor{easygreen}{\textbf{Basic Following}}, \textcolor{middleyellow}{\textbf{Advanced Following}}, and \textcolor{hardred}{\textbf{Designer Level Following}}. \textcolor{easygreen}{\textbf{Basic Following}} prompts are constructed by combining attributes/relations and objects from a single attribute group, while \textcolor{middleyellow}{\textbf{Advanced Following}} prompts involve cross-group composition, as well as two specialized dimensions—\textit{text rendering} and \textit{style control}—to evaluate text rendering and aesthetic coherence. Finally, 100 real-world \textcolor{hardred}{\textbf{Designer Level Following}} prompts are manually curated. The rightmost column reports the number of prompts per subclass, all of which are further expanded into long-form variants.}
\centering
\input{tables/attr_compose_table}
\label{tab:concept-compse}
\vspace{-2mm}
\end{table}

\textbf{New Dimensions}.
\label{sec:NewDimensions}
To extend evaluation beyond conventional capabilities, we introduce three additional dimensions: \textit{text rendering}, \textit{style control}, and \textit{designer-level prompts}.
(i) \textbf{Text rendering} evaluates a model’s ability to reproduce typographic elements, using prompts sourced from the \emph{Lex-Art} corpus \cite{zhao2025lexartrethinkingtextgeneration}.
(ii) \textbf{Style control} assesses the model’s capacity to follow high-level artistic directives, with prompts curated from leading AIGC communities.
(iii) \textbf{Designer-level prompts} involve complex instructions that incorporate practical constraints and domain-specific knowledge, collected through manual annotation.
The text rendering and style control dimensions are grouped into the \textbf{Advanced Following} set, while the designer-level prompts form the \textbf{Designer-Level Following} set.

\textbf{Length Augmentation}.
Finally, for each generated prompt, we construct a long-form variant by expanding the content through natural language paraphrasing and elaboration, while faithfully preserving its original semantics. 

\textbf{TestMini Subset}. 
Following the above pipeline, the full bench contains 5,000 prompts and 12,074 binary questions.  
Detailed statistics on question-list lengths and entity densities are provided in the \textbf{Appendix}. 
For efficient evaluation, we further construct a subset of \textbf{554 prompts} and \textbf{1446 binary questions}. 
Carefully sampled to maintain balanced category proportions, it achieves higher semantic uniqueness than the full benchmark as illustrated in Fig.~\ref{fig:other-bench-analyze}.

\subsection{Evaluation Method of TIIF‑Bench}
\label{sec:Evaluation Method of TIIF‑Bench}
To overcome the limitations of traditional expert scorers (e.g., CLIP and open-set object detectors) in evaluating generated images, we propose an attribute-specific, fine-grained evaluation protocol. Our method leverages the world knowledge encoded in large vision-language models (VLMs) to assess the alignment between textual instructions and generated images. Specifically, we employ a dedicated VLM-based evaluator, termed \textbf{TIIF Evaluator}, to perform attribute-level verification. The design and training of this evaluator are described in Sec.~\ref{sec:Ablation and Training of the VLM Evaluator}.

As illustrated in Fig.~\ref{fig: data-pipeline}, given an input prompt, we first extract its $N$ core concepts $C=\{c_i\}_{i=1}^{N}$ from the constructed concept pools. For each concept $c_i$, we employ an closed-source sota LLM to generate a corresponding binary question $q_i$, forming a set of evaluation questions $Q=\{q_i\}_{i=1}^{N}$ with ground-truth answers $A=\{\hat{a}_i\}_{i=1}^{N}$. The generated image, together with the questions $Q$, is then input to the \textbf{TIIF Evaluator}, which produces $N$ predicted answers. The final evaluation score $s$ for the generated image is computed as

\[
s = \frac{1}{N} \sum\nolimits_{i=1}^{N} \mathbb{I}[a_i=\hat{a}_i],
\]

where $\mathbb{I}[\cdot]$ denotes the indicator function and $a_i$ represents the evaluator’s prediction for question $q_i$. For \textbf{designer-level} prompts, which often encode complex real-world constraints and human priors, we manually construct a long list of tailored evaluation questions to ensure reliable assessment.

For \textbf{style control} dimension, we provide reference images covering common aspect ratios (16:9, 5:4, 1:1, 4:5, and 9:16). Reference images in the TestMini subset are generated using Nano-Banana-Pro, while those in the full benchmark are generated using Qwen-Image, followed by manual verification. During evaluation, the generated image is compared with the reference image of the closest aspect ratio using the average similarity of CSD~\cite{somepalli2024measuringstylesimilaritydiffusion} and DINOv3~\cite{dinov3}.

For \textbf{text rendering} evaluation, we propose a Global Normalized Edit Distance (GNED) metric. Let $P=\{p_1,\ldots,p_m\}$ denote the set of text tokens specified in the prompt and $G=\{g_1,\ldots,g_n\}$ the set of OCR-extracted tokens from the generated image. GNED computes the minimal character-level normalized edit distance (NED) between words in $P$ and $G$ using the Hungarian algorithm to obtain an optimal bipartite matching $\mathcal{M}$. A penalty term $|m-n|$ is added to account for unmatched words (e.g., missing or hallucinated text), and the total cost is normalized by $\max(m,n)$:

\[
\text{GNED}(P,G) = \left(\sum\nolimits_{(i,j)\in\mathcal{M}}\text{NED}(p_i,g_j) + |m-n|\right) / \max(m,n).
\]

\begin{figure}[t]
  \centering
  \includegraphics[width=\linewidth]{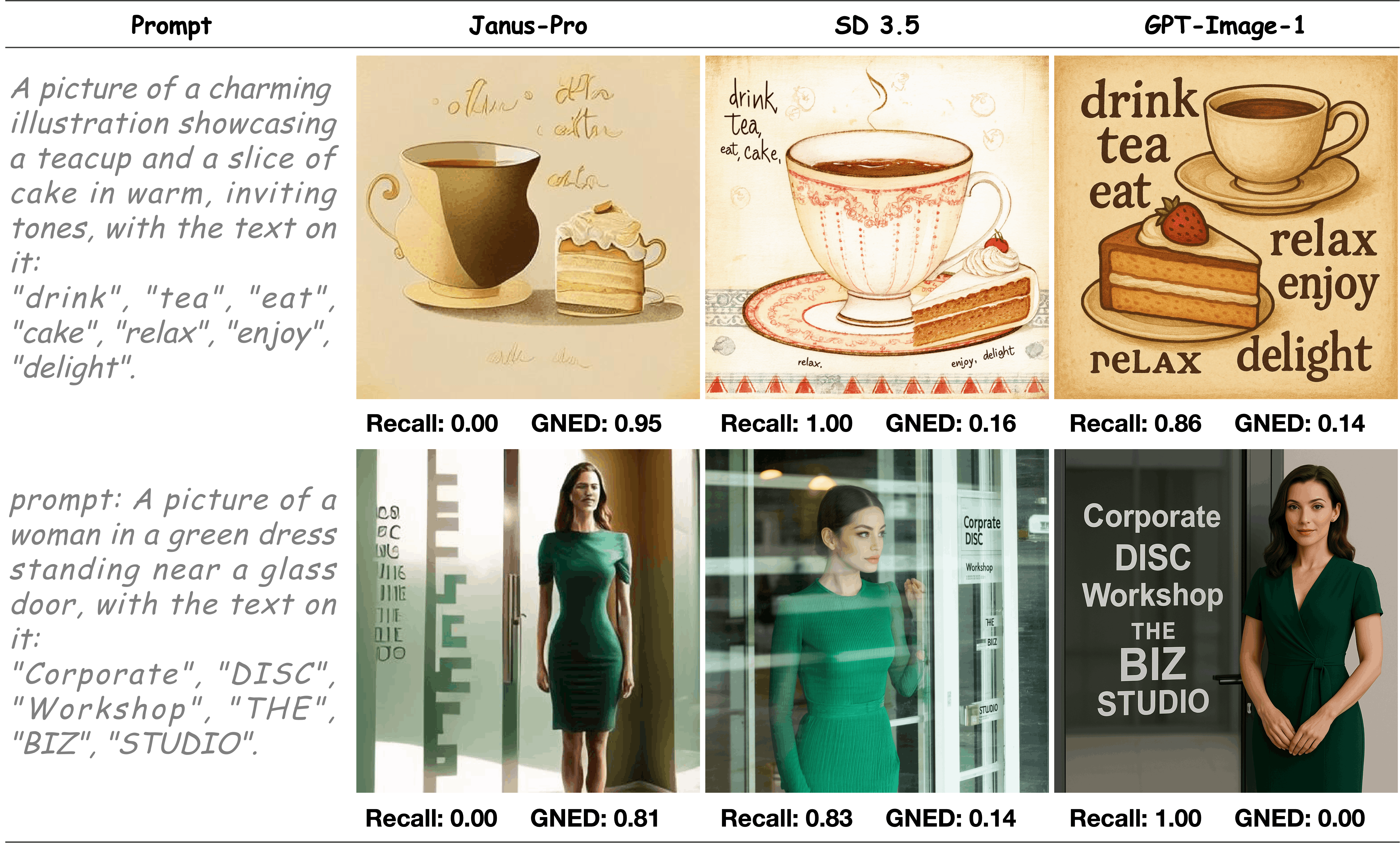}
  \caption{
    Visualization examples for Recall and GNED. 
    \textbf{Top row:} Janus-Pro\cite{chenJanusProUnifiedMultimodal} fails to generate any of the required words completely, resulting in a Recall of 0. Additionally, the overall quality of the generated text is poor, leading to a high GNED score. SD 3.5\cite{esserScalingRectifiedFlow2024} successfully generates all target words from the prompt, yielding a Recall of 1.00; however, the word “enjoy” contains minor distortions, resulting in a GNED of 0.16. GPT-Image-1\cite{ye2025echo4oharnessingpowergpt4o} omits the word “cake,” achieving a Recall of 6/7. Despite the omission, the visual quality of the generated words is higher than that of SD 3.5\cite{esserScalingRectifiedFlow2024}, thus resulting in a lower GNED score.  
    \textbf{Bottom row:} GPT-Image-1\cite{ye2025echo4oharnessingpowergpt4o} generates all required words with excellent typographic fidelity, achieving both the highest Recall score of 1.00 and the best GNED score of 0.00.
}
  \label{fig:recall-and-gned}
\end{figure}

GNED ranges from $0$ to $1$, where $0$ indicates perfect alignment and $1$ indicates maximal deviation. Compared with existing metrics such as OCR Recall and PNED~\cite{zhao2025lexartrethinkingtextgeneration}, GNED robustly penalizes both over-generation and omission, enabling consistent evaluation across samples with varying text lengths. Illustrative examples are shown in Fig.~\ref{fig:recall-and-gned}.

\begin{figure}[t]
  \centering
  \includegraphics[width=0.7\linewidth]{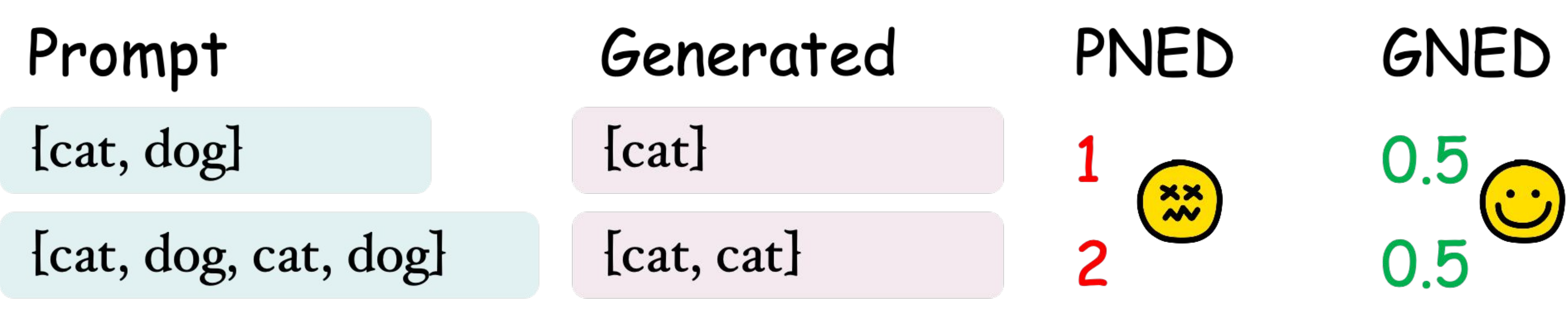}
  \caption{PNED can assign different scores to text-rendering errors with the same relative severity because its accumulated edit cost is not normalized across the complete target and prediction sets. GNED instead performs global matching and normalization, yielding scores that remain comparable across prompt lengths.}
  \label{fig:gned-pned-comparison}
\end{figure}

OCR Recall only measures whether target words are retrieved; it neither evaluates the rendering quality of each word nor penalizes hallucinated extra text. PNED can also assign different scores to errors with the same relative severity, as illustrated in Fig.~\ref{fig:gned-pned-comparison}. We further conducted a human study on 100 text-rendering cases sampled from short and long prompts and generated by Qwen-Image, GPT-Image-1, and NanoBanana. Using identical OCR outputs, OCR Recall, PNED, and GNED achieved 83\%, 94\%, and 99\% agreement with human pairwise preferences, respectively. GNED's advantage over PNED was concentrated on long prompts, where unnormalized accumulation tends to over-penalize partial omissions.

\subsection{Ablation and Training of TIIF Evaluator}
\label{sec:Ablation and Training of the VLM Evaluator}

To construct a diverse training corpus for the evaluator, we collect model outputs from \textbf{30 T2I models} spanning different release periods, architectures, generation capabilities, and resolutions. Each model is rolled out twice on all {5,000 prompts} in TIIF-Bench, producing approximately {300K generated images} and {724K binary questions}.
To build a reliable test set, we manually annotate {300 images}, resulting in {560 balanced questions} (280 “yes” and 280 “no”). This set is used exclusively for evaluator validation. On this test set, GPT-4o~\cite{hurst2024gpt} achieves an accuracy of \textbf{90.18\%}, indicating that it can serve as a reference evaluator. In contrast, the off-the-shelf Qwen3VL-8B~\cite{bai2025qwen3vltechnicalreport} reaches only \textbf{84.11\%}, suggesting that direct use of a general-purpose VLM may be insufficient for reliable evaluation.
We therefore use GPT-4o to annotate the remaining data (excluding the test set), during which chain-of-thought (CoT) reasoning is also generated. After filtering and balancing, we obtain a training set containing {109K positive} and {109K negative} binary questions.
We then investigate three training paradigms for Qwen3VL-8B:
\begin{enumerate}
\item \textbf{Direct prediction}: the model directly outputs the binary answer.
\item \textbf{Representation-based classification}: the final hidden state is extracted and fed into a two-layer MLP for binary classification supervision~\cite{zhang2025baserewardstrongbaselinemultimodal}.
\item \textbf{Reasoning-based prediction}: the model first generates chain-of-thought reasoning and then outputs the final answer.
\end{enumerate}

Each paradigm is trained for one epoch. On the held-out test set, direct prediction achieves \textbf{89.46\%} accuracy, while representation-based classification and reasoning-based prediction reach \textbf{91.43\%} and \textbf{91.07\%}. Both latter approaches meet the reliability requirement for serving as an automated evaluator.
Despite comparable accuracy, the reasoning-based paradigm provides interpretable chain-of-thought explanations that reveal where instruction-following failures occur. This additional interpretability is valuable for diagnosing model behavior and can potentially serve as reward signals for future model optimization. Therefore, we adopt the reasoning-based paradigm to construct the final \textbf{TIIF Evaluator}.

\section{Experiments}

\subsection{Performance of T2I Models on TIIF-Bench}
We benchmark TIIF-Bench on a diverse suite of widely used \emph{closed-source} and leading \emph{open-source} text-to-image models. 
The results are summarized in Tab.~\ref{tab: tiif_bench_results}. 
For each of the three difficulty levels, \emph{basic following}, \emph{advanced following}, and \emph{designer-level}. We report the average score of their corresponding dimensions under both short and long prompt settings. 
We further compute the overall average scores across all dimensions and difficulty levels. 
Note that GNED is a lower-is-better metric, the values reported in the \textit{Text} column are transformed as $1-\mathrm{GNED}$ so that higher scores consistently indicate better performance. 
From Tab.~\ref{tab: tiif_bench_results}, we derive several key observations, which are discussed below.

\begin{table}[t]
\captionsetup{skip=4pt}
\caption{Performance of closed-source models and state-of-the-art open-source models on TIIF-Bench. Evaluated systems are grouped into (i) \textcolor{objblue}{diffusion-based} open-source models, (ii) \textcolor{middleyellow}{unified} open-source models, and (iii) \textcolor{attrred}{closed-source} models; within each group, the highest score is shown in bold and color-coded for that group.}
\centering
\input{tables/tiif_bench_results}
\label{tab: tiif_bench_results}
\end{table}

\subsubsection{Diffusion-based Open-Source Models}

\textbf{(i) Overall performance}.
From the upper panel of Tab.~\ref{tab: tiif_bench_results}, we observe that Qwen-Image~\cite{wu2025qwenimagetechnicalreport} and FLUX.2 Dev~\cite{flux.2} exhibit the strongest instruction-following capability among diffusion-based open-source models. 
Closely following these two models are LuminaImage2~\cite{qin2025luminaimage20unifiedefficient} and FLUX.1 Dev~\cite{flux2024}. 
Overall, models built upon the MMDiT architecture continue to dominate the current landscape of high-quality image generation.
\textbf{(ii) Text Rendering}.
Among diffusion-based open-source models, only a subset explicitly supports in-image text rendering, including Qwen-Image, FLUX-1 Dev, FLUX-2 Dev, SANA 1.5~\cite{xieSANA15Efficient2025} (and SANA Sprint~\cite{chenSANASprintOneStepDiffusion2025}), and the Stable Diffusion family (SD-3~\cite{rombachHighResolutionImageSynthesis2022}, SD-3.5~\cite{esserScalingRectifiedFlow2024}). 
While SD-3 and SD-3.5 achieve competitive performance with short prompts, their effectiveness deteriorates when prompts become longer. This behavior is likely due to the increased difficulty of isolating textual elements from the broader descriptive context.
\textbf{(iii) Style Control}.
Prompt length has contrasting effects on different models. 
In systems such as PixArt-Alpha, the internal vocabulary for style control appears limited; for example, style tokens like \emph{Ghibli} or \emph{Cyberpunk} are weakly grounded in the learned representation. Consequently, short prompts that rely solely on such keywords often fail to produce coherent stylistic outputs. In contrast, longer prompts that provide richer visual descriptions can partially compensate for this limitation and significantly improve generation quality. 
Conversely, models such as SD-3.5 and PixArt-Sigma~\cite{chenPixArtSWeaktoStrongTraining2024} are typically trained on datasets where stylistic instructions appear as concise label-like tokens. When these style cues are embedded within longer descriptive prompts, their salience can be diluted, leading to reduced stylistic fidelity.
\textbf{(iv) Designer-Level Prompts}.
Designer-level prompts contain the densest and most diverse set of requirements, providing the most comprehensive evaluation of a model’s instruction-following capability. Consequently, the ranking of models under this setting closely mirrors the overall performance ranking.
\textbf{(v) Robustness to Prompt Length}.
Top-performing models demonstrate strong robustness to variations in prompt length, producing consistent results for both short and long versions of semantically equivalent prompts. In contrast, weaker models such as SD-3 and PixArt-Alpha exhibit substantial performance discrepancies between the two settings. This observation suggests that the ability to accurately interpret complex instructions is positively correlated with the overall image generation capability of a T2I model.

\subsubsection{Unified Open-Source Models}

\textbf{(i) Overall Performance}.
From the middle panel of Tab.~\ref{tab: tiif_bench_results}, we observe that the Emu3.5 series~\cite{cui2025emu35nativemultimodalmodels} achieves the strongest overall instruction-following performance among unified open-source models. This advantage likely stems from its large-scale pretraining and unified training paradigm that jointly optimizes image generation and multimodal understanding.
\textbf{(ii) Text Rendering}.
Autoregressive architectures are inherently less effective at rendering text within images. Except for Emu3.5, most unified models exhibit significantly weaker text-rendering performance compared with diffusion-based open-source state-of-the-art models of similar scale and release time.
\textbf{(iii) Style Control}.
Emu3.5 also demonstrates strong capability in style control. Its unified transformer backbone, trained jointly for generation and understanding, appears to facilitate more coherent global visual interpretation, which benefits style-conditioned generation.
\textbf{(iv) Designer-Level Prompts}.
Except for Emu3.5, most unified models struggle to handle complex real-world prompts at the designer level. Nevertheless, the ranking of models under this setting remains broadly consistent with their overall performance.
\textbf{(v) Robustness to Prompt Length}.
Stronger autoregressive models maintain relatively stable performance across both short and long prompt settings, whereas weaker models exhibit substantial performance gaps. This observation further supports the connection between robust instruction understanding and high-quality image generation.
\subsubsection{Diffusion-based vs. AR-based Open-Source Models}
\begin{figure}[t]
  \centering
  \includegraphics[width=\linewidth]{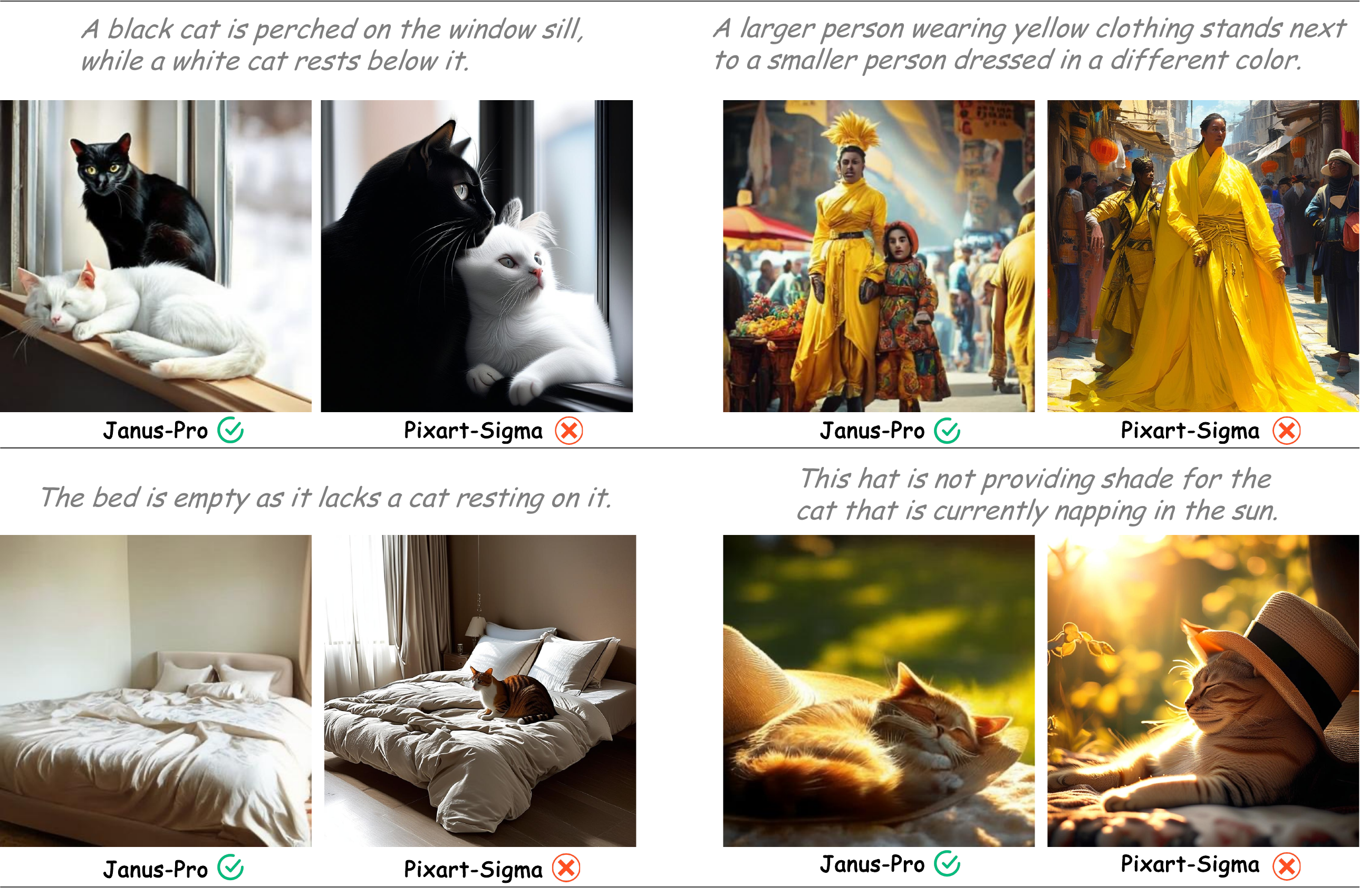}
  \caption{We observe that AR-based Janus-Pro outperforms the diffusion-based PixArt-Sigma on prompts requiring reasoning skills such as \textbf{differentiation}, \textbf{comparison}, and \textbf{negation}.}
  \label{fig: examples-ar-vs-diffusion}
\end{figure}

Although AR-based models tend to produce images with lower visual fidelity, their autoregressive architecture, jointly trained on generation and understanding tasks, grants them strong instruction-following capabilities. For instance, Janus-Pro\cite{chenJanusProUnifiedMultimodal} outperforms diffusion-based model PixArt-Sigma\cite{chenPixArtSWeaktoStrongTraining2024} on TIIF Bench. Representative examples are shown in Fig.~\ref{fig: examples-ar-vs-diffusion}.

\subsubsection{Closed-Source Models}
\textbf{(i) Overall Performance}.
From the bottom panel of Tab.~\ref{tab: tiif_bench_results}, we observe that the Nano-Banana series~\cite{nano-banana-pro,nano-banana} and GPT-Image-1~\cite{hurst2024gpt} demonstrate substantially stronger instruction-following capability than other models. This advantage likely stems from their unified architectures and superior language understanding. These models perform consistently well across multiple dimensions, including object attributes, spatial relations, and complex logical reasoning. In particular, they achieve exceptionally high success rates in text rendering, with no other model reaching comparable performance.
\textbf{(ii) Text Rendering}.
Benefiting from their strong multimodal comprehension capabilities, the Nano-Banana series~\cite{nano-banana-pro,nano-banana} and GPT-Image-1~\cite{hurst2024gpt} lead by a wide margin in the text rendering dimension, consistently producing accurate and legible in-image text.
\textbf{(iii) Style Control}.
With access to extensive knowledge and strong semantic understanding, the Nano-Banana series and GPT-Image-1 also dominate the style control, reliably generating images that align with the requested artistic styles.
\textbf{(iv) Designer-Level Prompts}.
Closed-source models generally perform well on designer-level prompts, which involve dense and diverse instruction requirements. This advantage likely reflects exposure to large-scale, high-quality training data. In particular, the Nano-Banana series and GPT-Image-1 show clear performance gains over open-source state-of-the-art models under this challenging setting.
\textbf{(v) Robustness to Prompt Length}.
Closed-source models exhibit strong robustness to variations in prompt length. Their performance remains stable across both short and long prompt settings, suggesting that well-trained multimodal architectures with large-scale data are better equipped to maintain consistent instruction understanding and generation quality.

\subsection{TIIF-Bench vs. Existing Benchmarks}
We compare TIIF-Bench with GenAI Bench and CompBench++ using four strong open-source models (SD 3.5, SANA 1.5, PixArt-Sigma, and Janus-Pro) and four representative closed-source models (GPT-Image-1, DALL-E 3, NanoBanana, and MidJourney v7). For every benchmark, all models are evaluated on the complete prompt set using the benchmark's official evaluation protocol.

\begin{table}[t]
\captionsetup{skip=4pt}
\caption{Performance of eight T2I models on {GenAI Bench}\cite{li2024genaibenchevaluatingimprovingcompositional}, evaluated using VQAScore. Results are reported for both basic and advanced prompt categories, as well as the overall average. }
\centering
\input{tables/genai-bench-8models}
\label{tab:genai-bench-8models}
\end{table}

\begin{table}[t]
\captionsetup{skip=4pt}
\caption{Evaluation results on {CompBench++}\cite{huang2025t2icompbenchenhancedcomprehensivebenchmark}. {CompBench++} provides both a GPT-based evaluation method and an expert model-based evaluation method; we report results from both.}
\centering
\input{tables/compbench-8models}
\label{tab:compbench-8models}
\end{table}

On GenAI Bench, VQAScore produces highly similar scores across models, limiting its ability to discriminate fine-grained instruction-following differences. On CompBench++, several models receive identical scores, and expert-based and GPT-based protocols can disagree substantially; for example, MidJourney v7 is rated highly by GPT-4o but markedly lower by expert models.

\begin{table}[t]
  \centering
  \captionsetup{skip=4pt} 
  \caption{Alignment of expert-based and GPT-based evaluations with human preference on \textsc{CompBench++}.}
  \label{tab:compbenchpp_Spearman}
  \input{tables/compbenchpp_Spearman}
\end{table}

To measure alignment with human judgment, we compute Spearman's rank correlation between benchmark-induced rankings and human rankings. CompBench++ shows weak or unstable alignment in several dimensions, especially when expert models fail on numeracy or spatial reasoning and when coarse GPT prompts induce hallucination.

\begin{table}[t]
\captionsetup{skip=4pt}
\caption{Alignment of TIIF-Bench evaluation results with human preference. GNED is used as the metric for the text rendering dimension, while all other dimensions are evaluated using VLM-based scoring.}
\centering
\input{tables/TIIF-Bench-Spearman}
\label{tab:TIIF-Bench-Spearman}
\end{table}

\begin{figure}[t]
  \centering
  \includegraphics[width=\linewidth]{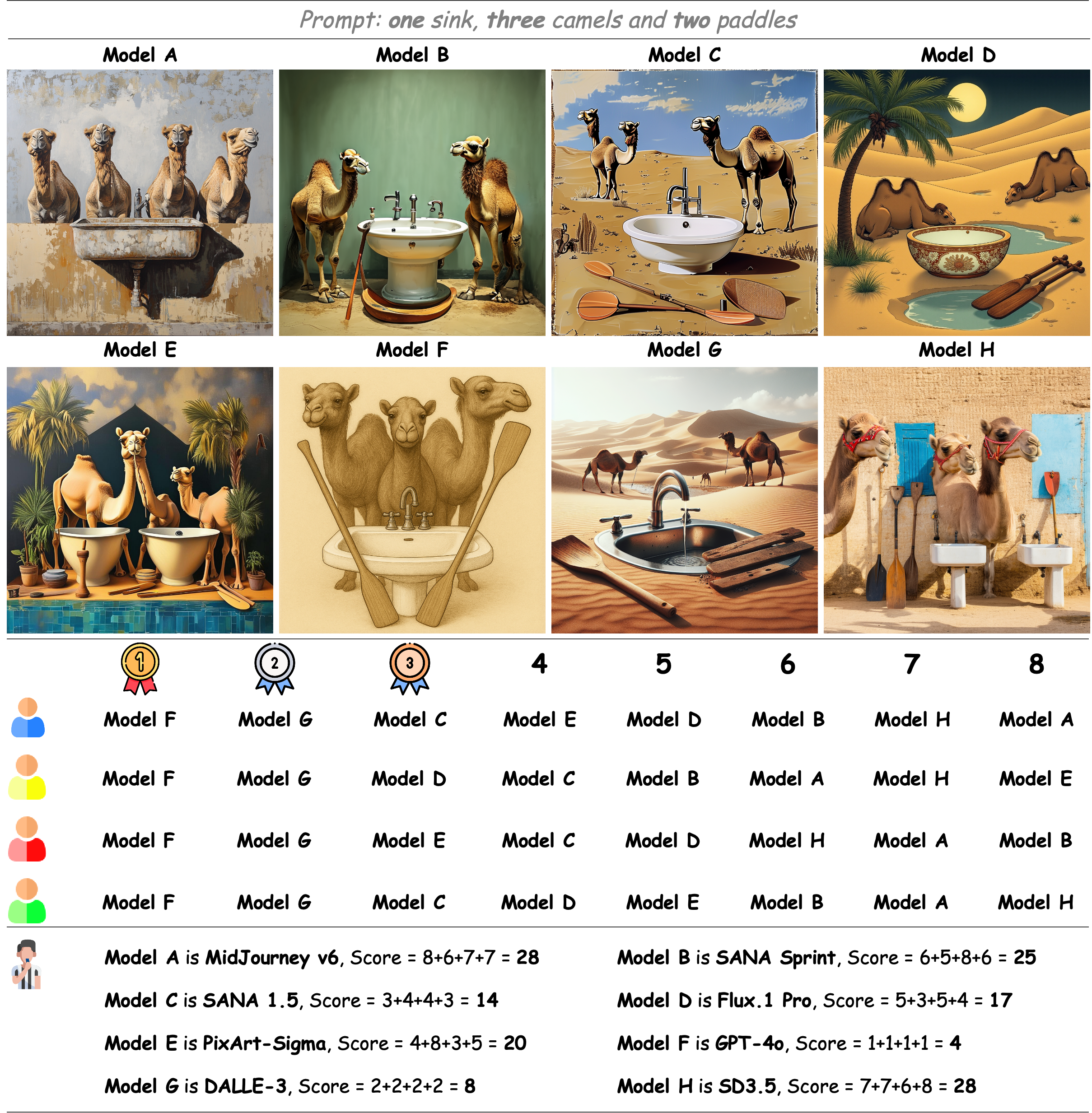}
  \caption{Illustration of the human preference study protocol. Each evaluation set contains one prompt from a specific dimension in \textsc{TIIF-Bench} and eight corresponding images, one generated by each candidate T2I model. Participants are asked to rank the eight images according to overall human preference, considering both instruction-following accuracy and overall visual quality. In practice, three such sets are sampled for each benchmark dimension, and the resulting rankings are aggregated to obtain a dimension-level human ranking.}
  \label{fig:spareman-example}
\end{figure}

We additionally conduct a blinded human preference study with ten independent participants. For each benchmark dimension, three prompt sets are sampled; each set contains eight images generated by the candidate models. Participants rank the images according to instruction-following accuracy and overall visual quality, and the rankings are aggregated into a dimension-level consensus. TIIF-Bench achieves consistently high agreement with these human preferences, supporting the reliability of its fine-grained evaluation protocol.

\section{Conclusion}
We introduce \textsc{TIIF-Bench}, a comprehensive and hierarchical benchmark for evaluating the instruction-following capabilities of text-to-image (T2I) models. 
TIIF-Bench covers diverse concept combinations and multiple evaluation dimensions, including text rendering, style control, and designer-level prompts. 
To enable reliable and fine-grained evaluation, we explore the paradigm of using VLMs as automated evaluators and develop a reproducible reasoning-based evaluator, termed \textbf{TIIF Evaluator}. 
We further propose GNED for evaluating text rendering quality, and provide high-quality aspect-ratio-diverse reference images for style control prompts. 
Extensive experiments on TIIF-Bench reveal key patterns in how current T2I models follow complex instructions during image generation, offering useful insights for future T2I research.

\textbf{Limitations}. Despite TIIF-Bench's advantages over existing benchmarks, there remain some limitations. First, the prompts contain mainly common objects. In future work, we will include obscure vocabulary or rare objects to further challenge the instruction following capability of T2I models. Second, currently, all prompts are in English. Incorporating linguistic diversity is an important direction for future work. Last but not least, how could stylistic variations (\textit{e.g.}, formal vs. conversational) impact model performance is not considered yet. An investigation on this problem may offer valuable insight in practical applications. 

%
%
\renewcommand{\thesection}{A\arabic{section}}
\renewcommand{\thesubsection}{A\arabic{section}.\arabic{subsection}}
\renewcommand{\thefigure}{A\arabic{figure}}
\renewcommand{\thetable}{A\arabic{table}}
\setcounter{section}{0}
\setcounter{figure}{0}
\setcounter{table}{0}
\renewcommand{\theHsection}{appendix.\arabic{section}}
\renewcommand{\theHsubsection}{appendix.\arabic{section}.\arabic{subsection}}
\renewcommand{\theHfigure}{appendix.\arabic{figure}}
\renewcommand{\theHtable}{appendix.\arabic{table}}

\section{Benchmark Statistics}

\subsection{Details of the Concept Pools}
\label{sec:details_of_concept_pools}

\begin{table}[t]
\captionsetup{skip=4pt}
\caption{The ten dimensions used to construct the TIIF-Bench prompt set.
We first mine these dimensions from existing benchmarks and regroup them into three broad categories: \textit{Attributes}, \textit{Relations}, and \textit{Reasoning}.
For concepts extracted from \textsc{CompBench++}, we leverage GPT-4o~\cite{hurst2024gpt} to separate the core \textcolor{objblue}{objects} from their associated \textcolor{attrred}{attributes/relations}; the two components are then stored independently for subsequent composition.
In contrast, concepts derived from \textsc{GenAI Bench} are retained as complete phrases, since their action dependencies and logical structures---such as \emph{differentiation}, \emph{comparison}, and \emph{negation}---cannot be cleanly factorized without altering the original semantics.}
\centering
\input{tables/attr_pools_table}
\label{tab:attribute-pools}
\end{table}

Table~\ref{tab:attribute-pools} summarizes the ten dimensions used to construct TIIF-Bench, which we organize into three broad categories: \textit{Attributes}, \textit{Relations}, and \textit{Reasoning}. This taxonomy is designed to cover progressively more challenging forms of instruction following, ranging from local attribute binding to relational understanding and higher-level reasoning.

The concept pools are mined from existing T2I benchmarks and then reorganized into a reusable composition space. In particular, concepts from \textsc{CompBench++} are decomposed into object-centric units and their associated attributes or relations, which enables flexible recombination during prompt construction. By contrast, prompts from \textsc{GenAI Bench} often encode tightly coupled semantic structures, such as comparison, negation, and differentiation, and are therefore preserved as whole phrases rather than being decomposed into independent components.

This design allows TIIF-Bench to balance controllable prompt composition with semantic fidelity to the original benchmark sources, thereby supporting both diversity and interpretability in the resulting prompt set.

\subsection{Prompt Length and Checklist Statistics}
\label{sec:prompt_checklist_stats}

To further characterize the composition of TIIF-Bench, we report the average prompt length and the average checklist length for each evaluation dimension in Table~\ref{tab:stats}. Prompt length is measured in words, and checklist length denotes the average number of binary verification questions associated with each prompt. Since the short and long versions of a prompt share the same semantic target, they are evaluated using the same checklist; therefore, only one checklist statistic is reported for each dimension.

\begin{table}[t]
\captionsetup{skip=4pt}
\caption{Statistics of prompt length and checklist length across different dimensions of TIIF-Bench. ``Avg. Len.'' denotes the average prompt length in words, and ``Check Len.'' denotes the average number of binary questions in the corresponding checklist. Short and long prompts share the same checklist.}
\centering
\input{tables/stats}
\label{tab:stats}
\end{table}

Overall, these statistics highlight two core properties of TIIF-Bench: diversity in prompt formulation and variability in evaluation granularity. Together, they make the benchmark better suited for diagnosing instruction-following behavior under both linguistic variation and task-specific complexity.

\section{Meta Prompts}
\label{sec:Meta Prompts}
\subsection{Meta Prompt for Length Augmentation}
For each generated prompt, we use GPT-4o\cite{hurst2024gpt} to create a longer version of it by paraphrasing and elaborating in natural language, while preserving the original meaning. The prompt is as follows:

\begin{tcolorbox}[colback=gray!5, colframe=black!30, boxrule=0.4pt,
                 arc=2mm, left=4pt, right=4pt, top=4pt, bottom=4pt,
                 fontupper=\small\ttfamily]
\#\#\#RAW CAP\#\#\#

You are a professional writer. Observe the sentence above, which describes a visual scene.\\
Please expand the sentence by increasing its linguistic richness. You may elaborate with more complex structure, rhetorical flourishes, or stylistic details—however, **DO NOT** introduce any new objects, entities, or events. The visual content must remain unchanged.
\\Return only the expanded sentence, without any extra explanation or commentary.
\end{tcolorbox}

\subsection{Meta Prompt for Evaluation}
For each prompt and its corresponding generated image, we insert the yes/no questions into a predefined meta prompt. The resulting prompt, along with the image, is then passed to TIIF-Evaluator for evaluation. The meta prompt is as follows:
\begin{tcolorbox}[colback=gray!5, colframe=black!30, boxrule=0.4pt,
                 arc=2mm, left=4pt, right=4pt, top=4pt, bottom=4pt,
                 fontupper=\small\ttfamily]
You are a careful image understanding assistant.\\

You will receive an image and a binary question. Your job is to answer the question using only the visual content of the image.\\

Rules:\\
1. Base your judgment strictly on what is visible in the image.\\
2. Do not assume facts that are not visually supported.\\
3. Pay attention to objects, attributes, counts, spatial relations, actions, and negation if relevant.\\
4. If the image is ambiguous or the evidence is weak, explain that in the reasoning, but still provide the most likely binary answer.\\
5. The final answer must be exactly one of: Yes, No.\\
6. First provide reasoning, then provide the final answer.\\

Use exactly this format:\\
CoT: <step-by-step visual reasoning grounded in the image>\\
Answer: <Yes or No>
\end{tcolorbox}

\section{Additional Experimental Results}
\label{sec:More Experiment Results}
\subsection{Human Preference Study}
\label{sec:user_study}

The main paper reports the complete human preference protocol and correlation results.


\subsection{Additional Experiments on Text Rendering} 

We conduct additional evaluations on the text rendering capability of T2I models with these metrics. The results are summarized in Tab.~\ref{tab: text_gen_results}.

\begin{table}[t]
\captionsetup{skip=4pt}
\centering
\caption{Performance of open-source and closed-source models with text rendering capabilities. GNED and Recall are used as quantitative evaluation metrics.}
\begin{minipage}{0.8\linewidth}
    \centering
    \centering
    \input{tables/text_gen_results}
    \label{tab: text_gen_results}
\end{minipage}
\end{table}

\begin{figure}[t]
  \centering
  \begin{minipage}{0.999\linewidth}
    \centering
    \includegraphics[width=\linewidth]{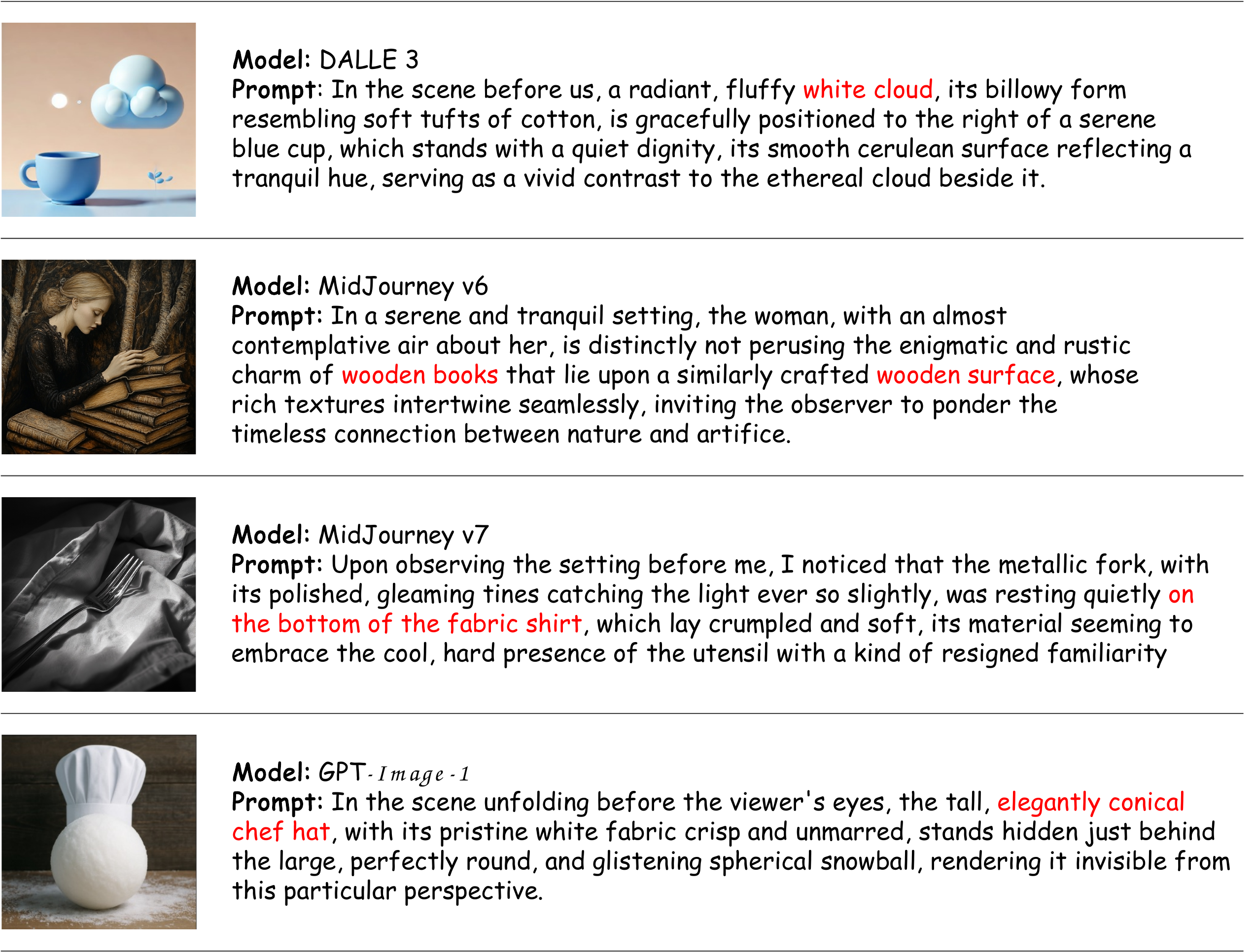}
    \caption{Failure cases of proprietary models on prompts involving spatial relations.}
    \label{fig:examples-poor-relation}
  \end{minipage}
\end{figure}

\subsection{Training Details and Additional Validation of TIIF-Evaluator}
\begin{figure}[t]
  \centering
  \begin{minipage}{0.999\linewidth}
    \centering
    \includegraphics[width=\linewidth]{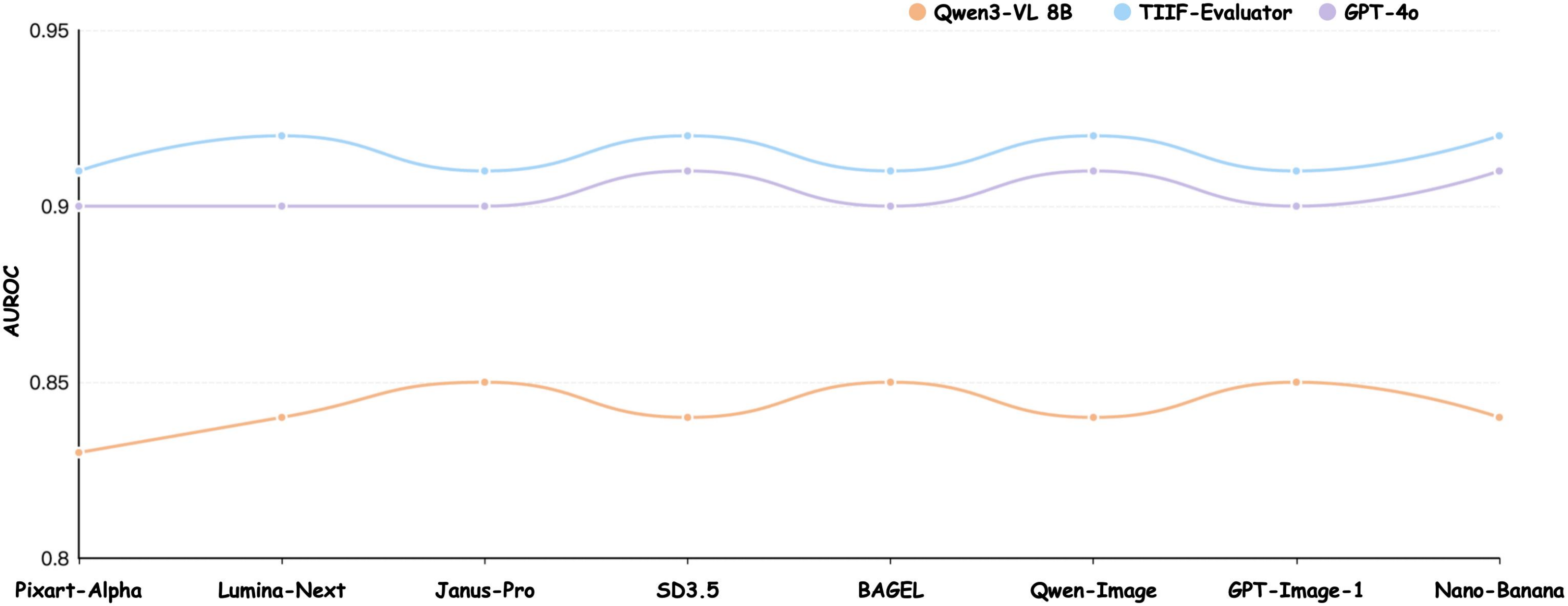}
    \caption{AUROC of different evaluators against human checklist annotations across eight representative T2I models: PixArt-Alpha\cite{chenPixArt$a$FastTraining2023}, Lumina-Next\cite{zhuoLuminaNextMakingLuminaT2X2024}, Janus-Pro\cite{chenJanusProUnifiedMultimodal}, SD3.5\cite{esserScalingRectifiedFlow2024}, BAGEL\cite{deng2025emergingpropertiesunifiedmultimodal}, Qwen-Image\cite{wu2025qwenimagetechnicalreport}, GPT-Image-1\cite{hurst2024gpt}, and Nano-Banana\cite{nano-banana}. For each model, we sample one prompt from each of the nine dimensions in \textsc{TIIF-Bench}, generate images, and collect human answers to all checklist questions. The trained \textsc{TIIF-Evaluator} consistently achieves the highest agreement with human labels, followed by GPT-4o\cite{hurst2024gpt}, while the off-the-shelf Qwen3VL-8B\cite{bai2025qwen3vltechnicalreport} clearly weaker.}
    \label{fig:auroc}
  \end{minipage}
\end{figure}

To enable scalable and fine-grained evaluation on \textsc{TIIF-Bench}, we develop a dedicated VLM-based evaluator, termed \textsc{TIIF-Evaluator}. Unlike existing benchmark pipelines that either rely on weak expert scorers or directly adopt a general-purpose VLM without task-specific adaptation, our evaluator is explicitly trained for attribute-level binary verification under the \textsc{TIIF-Bench} protocol.

\paragraph{Training corpus construction.}
To construct a diverse training corpus for the evaluator, we collect model outputs from 30 T2I models spanning different release periods, model architectures, generation capabilities, and output resolutions. Each model is rolled out twice on all 5,000 prompts in \textsc{TIIF-Bench}, resulting in approximately 300K generated images and 724K binary questions. This corpus exposes the evaluator to a broad range of success and failure cases across heterogeneous T2I systems.

\paragraph{Held-out validation set.}
To build a reliable test set for evaluator selection, we manually annotate 300 generated images and obtain 560 balanced binary questions, including 280 positive (\textit{yes}) and 280 negative (\textit{no}) labels. This set is used exclusively for evaluator validation and is excluded from the evaluator training data.

\paragraph{Reference annotation and paradigm comparison.}
We first evaluate GPT-4o\cite{hurst2024gpt} and the off-the-shelf Qwen3VL-8B\cite{bai2025qwen3vltechnicalreport} on the held-out validation set. GPT-4o\cite{hurst2024gpt} achieves 90.18\% accuracy, while the off-the-shelf Qwen3VL-8B\cite{bai2025qwen3vltechnicalreport} reaches only 84.11\%, indicating that direct use of a general-purpose VLM is insufficient for reliable fine-grained evaluation. We therefore use GPT-4o\cite{hurst2024gpt} as a reference annotator for the remaining data outside the held-out set. During this process, GPT-4o\cite{hurst2024gpt} also produces chain-of-thought (CoT) reasoning together with the final binary answer. After filtering and balancing, we obtain a final training set containing 109K positive and 109K negative binary questions.

Based on this training set, we investigate three training paradigms for Qwen3VL-8B:
(1) \textit{Direct prediction}, where the model directly outputs the binary answer;
(2) \textit{Representation-based classification}, where the final hidden state is extracted and fed into a two-layer MLP for binary classification supervision\cite{zhang2025baserewardstrongbaselinemultimodal};
and (3) \textit{Reasoning-based prediction}, where the model first generates CoT reasoning and then outputs the final answer.
Each paradigm is trained for one epoch. On the held-out validation set, direct prediction achieves 89.46\% accuracy, while representation-based classification and reasoning-based prediction reach 91.43\% and 91.07\%, respectively. Although the latter two paradigms achieve comparable performance, we adopt reasoning-based prediction as the final \textsc{TIIF-Evaluator}, since it provides interpretable reasoning that helps reveal where instruction-following failures occur.

\paragraph{Additional validation across different T2I models.}
To further assess evaluator reliability under different image-generation characteristics, we conduct an additional validation experiment on eight representative T2I models covering diverse generation quality and instruction-following capability: \textit{PixArt-Alpha}, \textit{Lumina-Next}, \textit{Janus-Pro}, \textit{SD3.5}, \textit{BAGEL}, \textit{Qwen-Image}, \textit{GPT-Image-1}, and \textit{Nano-Banana}. For each model, we sample one prompt from each of the nine dimensions in \textsc{TIIF-Bench}, generate the corresponding images, and ask professional annotators to answer all checklist questions manually, thereby obtaining human labels.

We then compare three evaluators---the off-the-shelf \textit{Qwen3VL-8B}\cite{bai2025qwen3vltechnicalreport}, the trained \textsc{TIIF-Evaluator}, and \textit{GPT-4o}\cite{hurst2024gpt}---against these human annotations, and compute the AUROC for each model separately. The results are shown in Fig.~\ref{fig:auroc}. 
The off-the-shelf Qwen3VL-8B\cite{bai2025qwen3vltechnicalreport} consistently underperforms the other two evaluators, further confirming that direct deployment of a general-purpose VLM is not sufficiently reliable for fine-grained binary verification on \textsc{TIIF-Bench}. 
\textsc{TIIF-Evaluator} consistently achieves the highest AUROC among the three evaluators. This finding complements the held-out accuracy reported in the main paper and further supports our choice of the reasoning-based paradigm for the final evaluator.

Overall, these results indicate that the advantage of \textsc{TIIF-Evaluator} is not limited to improved average accuracy on a single held-out set. Rather, it remains better aligned with human judgments across a diverse collection of T2I systems with different visual characteristics and generation behaviors, which is essential for a benchmark designed to compare heterogeneous models in a fair and fine-grained manner.

\section{Additional Visualizations}
\label{sec:Additional Visualizations}
Unless otherwise specified, the main benchmark results and the human preference study use the latest available model versions at the time of evaluation. In particular, MidJourney v7\cite{Midjourney} is used in the main benchmark tables and the human study, while some qualitative examples in this appendix additionally include MidJourney v6 outputs for visual comparison and case analysis.

\subsection{Failure Cases of Strong Models in the Relation Dimension}
\label{sec: Failure Cases of Strong Closed-source Models in the Relation Dimension}
Through extensive experiments on our \textsc{TIIF-Bench}, we observe that most models exhibit strong instruction-following capabilities in object attribute dimensions such as color and material; however, their performance degrades when handling spatial relations. Fig.~\ref{fig:examples-poor-relation} presents several failure cases from strong closed-source models when generating images for prompts involving spatial relations.

\subsection{Additional Examples of Style Control}
\label{sec:Additional Examples of Style Control}

Style control is introduced in \textsc{TIIF-Bench} as a dedicated evaluation dimension for measuring whether a T2I model can follow high-level artistic directives while preserving semantic consistency with the input prompt. Unlike conventional compositional prompts that mainly test local object attributes or relations, style-control prompts require the model to coordinate multiple global factors simultaneously, including rendering style, color palette, texture, composition, and overall aesthetic coherence. This makes style control a substantially different challenge from standard instruction following and an important capability in practical creative applications.

To support reliable evaluation on this dimension, each style-control prompt in \textsc{TIIF-Bench} is paired with a set of reference images covering several common aspect ratios, including 16:9, 5:4, 1:1, 4:5, and 9:16. During evaluation, the generated image is compared with the reference image of the closest aspect ratio, which reduces incidental mismatches caused by layout differences and makes the score more reflective of actual stylistic fidelity. Following the protocol used in the main paper, the final style-control score is computed as the arithmetic mean of three similarity measures: CSD\cite{somepalli2024measuringstylesimilaritydiffusion}, DINOv3\cite{dinov3}, and SigLIP2\cite{tschannen2025siglip2multilingualvisionlanguage}. This reference-based design allows the benchmark to assess whether a model captures the intended global style in a visually coherent manner, rather than merely responding to a few isolated style keywords.

\begin{figure}[t]
  \centering
  \includegraphics[width=0.9\linewidth]{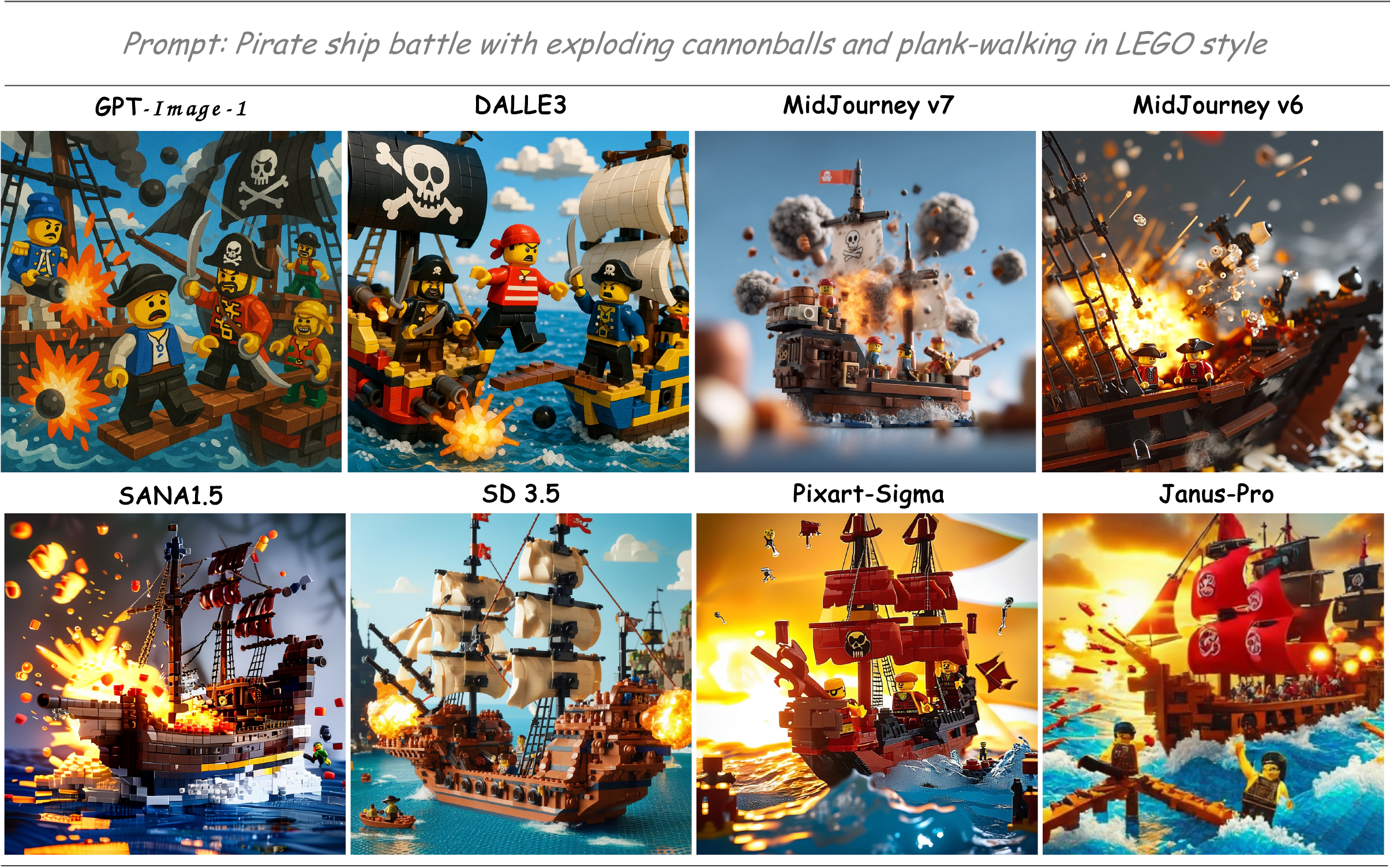}
  \captionof{figure}{Representative examples from the \textit{Style Control} dimension of \textsc{TIIF-Bench}. 
This dimension evaluates whether a T2I model can follow high-level artistic directives while maintaining semantic correctness and global aesthetic coherence. 
Each example reveals clear differences in how current models capture the intended style, visual texture, color composition, and overall rendering quality.}
  \label{fig: examples-style-control}
\end{figure}

Figure~\ref{fig: examples-style-control} presents representative examples from this dimension. Several characteristic patterns can be observed. First, strong models are often able to reproduce the target visual identity at the global level, including the intended artistic medium, dominant color distribution, scene atmosphere, and rendering texture. Second, weaker models frequently exhibit only partial style adherence: although they may preserve the coarse semantic content of the prompt, the generated image can still deviate substantially from the desired visual language, compositional mood, or fine-grained stylistic details. Third, style control is notably sensitive to prompt formulation. As discussed in the main paper, some models benefit from longer prompts that provide richer stylistic descriptions, whereas others perform better when style cues are expressed in short, label-like forms. The examples in Fig.~\ref{fig: examples-style-control} qualitatively illustrate these differences and complement the quantitative results reported in the main paper.

Overall, this dimension highlights that high-quality T2I generation should not be evaluated solely through object correctness or local image-text alignment. A strong model must also preserve the global stylistic intent of the prompt in a coherent and controllable manner, which is precisely the ability that the style-control dimension of \textsc{TIIF-Bench} is designed to probe.

\subsection{Additional Examples of Text Rendering}
\label{sec:Additional Examples of text rendering}
\begin{figure}[t]
  \centering
  \includegraphics[width=0.9\linewidth]{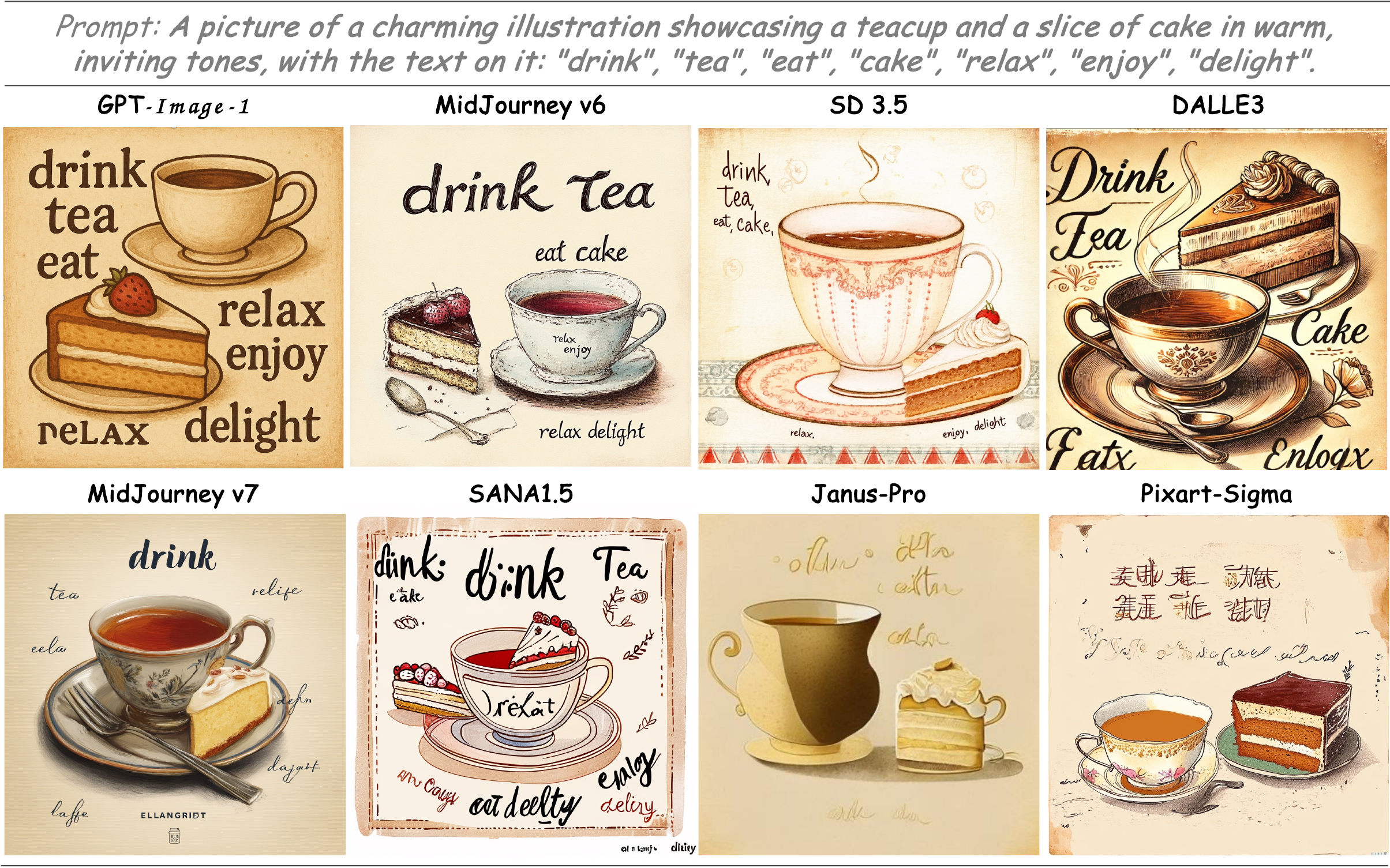}
  \caption{\textbf{Text Rendering} is introduced as a novel evaluation dimension for assessing a model’s ability to generate complex, non-natural textures—embedded human language text. We present illustrative examples for eight representative T2I models.}
  \label{fig: examples-text}
\end{figure}

\textsc{TIIF-Bench} introduces text rendering as a novel evaluation dimension to assess a model’s ability to generate complex, non-natural textures such as embedded text. We adopt two metrics for this task: OCR Recall and the proposed GNED. The main paper visualizes both metrics. In addition, Fig.\ref{fig: examples-text} presents qualitative examples on how different T2I models perform on this dimension.

\begin{figure}[t]
  \centering
  \includegraphics[width=0.9\linewidth]{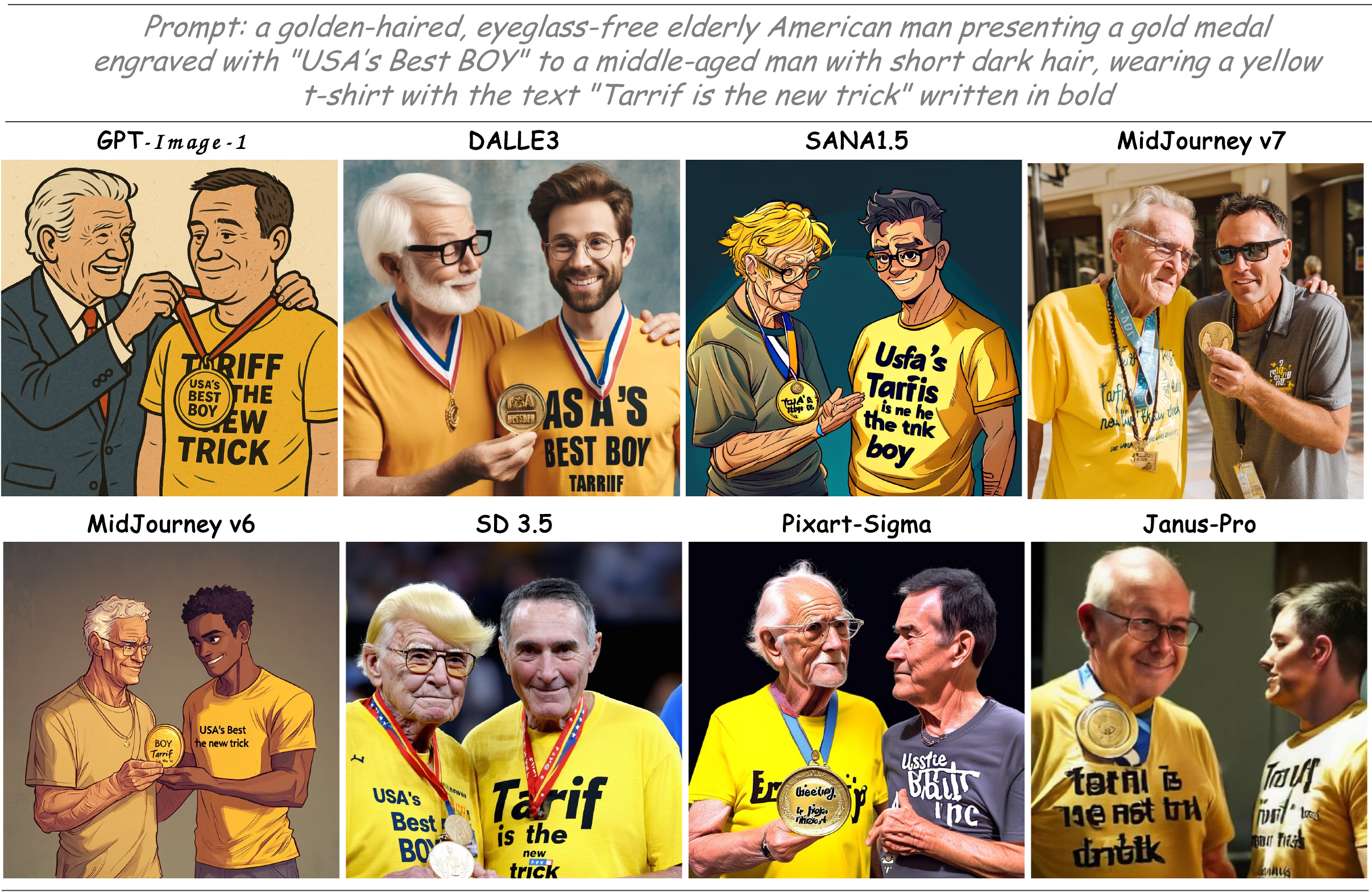}
  \caption{\textbf{Designer-level} prompts comprise complex and diverse requirements, offering the most comprehensive test of a model’s instruction-following capabilities. We provide illustrative examples for eight representative T2I models.}
  \label{fig: examples-real-world}
\end{figure}

\subsection{Additional Examples of Real-world Prompts}
\label{sec:Additional Examples of Real-world Prompts}
The designer-level prompts feature dense and diverse requirements, providing the most comprehensive evaluation of a model’s instruction-following capabilities. Fig.~\ref{fig: examples-real-world} shows examples on how current T2I models perform on this dimension.

\subsection{AR-based versus Diffusion-based Models}
As discussed in Sec.~4.1.3 of the main paper, although AR-based models typically generate images with lower visual fidelity, their autoregressive architecture—jointly trained on both generation and understanding tasks—enables strong instruction-following capabilities. 
The main paper provides the corresponding visual comparison. Janus-Pro outperforms diffusion-based model PixArt-Sigma on \textsc{TIIF-Bench}, especially on prompts involving reasoning logic such as differentiation, comparison, and negation.


\clearpage
\bibliographystyle{splncs04}
\bibliography{main}
\end{document}

%% file: tables/attr_compose_table.tex
\newcommand{\thinrule}{\textcolor{black!60}{\vrule width 0.03pt}}
\newcolumntype{!}{@{\hspace{4pt}\thinrule\hspace{2pt}}}
\renewcommand{\arraystretch}{1.3}   
\setlength{\tabcolsep}{4pt}

\small
\begin{adjustbox}{width=\linewidth}
\begin{tabular}{
  >{\raggedright\arraybackslash}m{1.8cm}!   
  >{\centering\arraybackslash}m{2.6cm}!      
  >{\raggedright\arraybackslash}p{9.4cm}!    
  >{\centering\arraybackslash}m{2.5cm}       
}
\toprule
\multicolumn{1}{c!}{\textbf{Level}} &
\multicolumn{1}{c!}{\textbf{Dimension}} &
\multicolumn{1}{c!}{\textbf{Combination Policy}} &
\multicolumn{1}{c}{\textbf{Count}} \\
\midrule
\multirow[c]{3}{*}{\makecell[l]{\textcolor{easygreen}{\textbf{Basic}}\\\textcolor{easygreen}{\textbf{Following}}}}
 & Attribute & \makecell[l]{Shape+Color,\; Color+Texture,\; Texture+Color,\; Shape+Texture} & \{100,50,50,100\}\\
 & Relation  & \makecell[l]{2D Spatial,\; 3D Perspective,\; Action+2D,\; Action+3D}         & \{75,75,75,75\} \\
 & Reasoning & Numeracy,\; Negation,\; Differentiation,\; Comparison                        & \{75,75,75,75\} \\
\midrule
\multirow[c]{5}{*}{\makecell[l]{\textcolor{middleyellow}{\textbf{Advanced}}\\\textcolor{middleyellow}{\textbf{Following}}}}
 & \makecell[c]{Attribute\\+ Relation}
 & \makecell[l]{Action+Color,\; Action+Texture,\; Color+2D,\; Color+3D,\\ Shape+2D,\; Shape+3D,\; Texture+2D,\; Texture+3D}
 & \makecell{$\{50,50,33,33,$\\$33,33,33,33\}$} \\[8pt]   

 & \makecell[c]{Attribute\\+ Reasoning}
 & \makecell[l]{Numeracy+\{Color, Texture\},\; Comparision+\{Color, Texture\},\\ Differentiation+\{Color, Texture\},\; Negation+\{Color, Texture\}}
 & \makecell{$\{40,40,40,40,$\\$40,40,40,40\}$} \\[8pt]   

 & \makecell[c]{Relation\\+ Reasoning}
 & \makecell[l]{Numeracy+\{2D, 3D\},\; Comparision+\{2D, 3D\},\\ Differentiation+\{2D, 3D\},\; Negation+\{2D, 3D\}}
 & \makecell{$\{40,40,40,40,$\\$40,40,40,40\}$} \\

 & Text Generation & — & \{150\} \\
 & Style Control   & — & \{150\} \\
\midrule
\makecell[l]{\rule{0pt}{3.5ex}\textcolor{hardred}{\textbf{Designer}}\\\textcolor{hardred}{\textbf{Level}}} & Complex & — & \{100\} \\
\bottomrule
\end{tabular}
\end{adjustbox}

%% file: tables/tiif_bench_results.tex
\newcommand{\thinrule}{\textcolor{black!60}{\vrule width 0.03pt}}
\newcolumntype{!}{@{\hspace{4pt}\thinrule\hspace{4pt}}}
\renewcommand{\arraystretch}{1.7} 
\setlength{\tabcolsep}{3pt}

\small
\centering
\begin{adjustbox}{width=\linewidth}
\begin{tabular}{
  >{\raggedright\arraybackslash}m{2.4cm}!           
  *{2}{>{\centering\arraybackslash}m{0.7cm}!}       
  *{8}{>{\centering\arraybackslash}m{0.65cm}!}      
  *{12}{>{\centering\arraybackslash}m{0.65cm}!}     
  >{\centering\arraybackslash}m{0.65cm}!            
  >{\centering\arraybackslash}m{0.65cm}             
}
\toprule
\multirow{3}{*}{\textbf{Model}}
  & \multicolumn{2}{c!}{\multirow{2}{*}{\textbf{Overall}}}
  & \multicolumn{8}{c!}{\textbf{Basic Following}}
  & \multicolumn{12}{c!}{\textbf{Advanced Following}}
  & \multicolumn{2}{c}{\textbf{Designer}} \\

\cmidrule(lr{.3em}){4-11} \cmidrule(lr{.3em}){12-23} \cmidrule(l){24-25}

& & &
  \multicolumn{2}{>{\centering\arraybackslash}m{1.20cm}!}{Avg}
  & \multicolumn{2}{>{\centering\arraybackslash}m{1.20cm}!}{\hspace{-4pt}Attribute}
  & \multicolumn{2}{>{\centering\arraybackslash}m{1.20cm}!}{\hspace{-4pt}Relation}
  & \multicolumn{2}{>{\centering\arraybackslash}m{1.20cm}!}{\hspace{-4pt}Reason}
  & \multicolumn{2}{>{\centering\arraybackslash}m{1.20cm}!}{Avg}
  & \multicolumn{2}{>{\centering\arraybackslash}m{1.20cm}!}{\makecell[c]{Attr.\\+Rela.}}
  & \multicolumn{2}{>{\centering\arraybackslash}m{1.20cm}!}{\makecell[c]{Attr.\\+Rea.}}
  & \multicolumn{2}{>{\centering\arraybackslash}m{1.20cm}!}{\makecell[c]{Rela.\\+Rea.}}
  & \multicolumn{2}{>{\centering\arraybackslash}m{1.20cm}!}{Style}
  & \multicolumn{2}{>{\centering\arraybackslash}m{1.20cm}!}{Text}
  & \multicolumn{2}{>{\centering\arraybackslash}m{1.20cm}}{\makecell[c]{Real\\World}} \\

\cmidrule(lr{.3em}){4-11} \cmidrule(lr{.3em}){12-23} \cmidrule(l){24-25}

& short & long &          
  short & long &          
  short & long &          
  short & long &          
  short & long &          
  short & long &          
  short & long &          
  short & long &          
  short & long &          
  short & long &          
  short & long &          
  short & long            
\\

\midrule
\addlinespace[-1pt]     
\multicolumn{23}{c}{\textbf{\textcolor{objblue}{Diffusion based} Open-Source Models}} \\[-1pt]
\midrule
Qwen-Image  &87.0 &87.5 &87.1 &87.5 &87.3 &88.0 &90.2 &90.8 &83.9 &84.5 &86.9 &87.6 &83.2 &83.8 &80.4 &81.2 &83.1 &83.7 &96.7 &97.2 &79.0 &81.0 &89.9 &90.6 \\
FLUX.2 dev  &87.2 &87.3 &86.6 &86.1 &87.3 &87.1 &90.0 &90.1 &82.5 &81.9 &89.3 &88.9 &85.5 &85.4 &90.5 &89.8 &81.8 &81.6 &100.0 &99.5 &71.0 &73.0 &91.3 &90.9 \\
LuminaImage2&71.2 &70.7 &79.7 &78.9 &74.7 &74.4 &85.0 &85.1 &79.0 &78.4 &66.9 &66.0 &78.6 &78.2 &75.9 &75.7 &70.7 &70.0 &86.7 &86.6 &20.0 &19.0 &66.7 &66.2 \\
FLUX.1 dev  &72.6 &71.0 &82.8 &80.6 &86.9 &82.5 &88.8 &81.5 &74.4 &73.9 &65.1 &69.6 &68.2 &74.4 &73.1 &72.7 &70.3 &70.9 &67.5 &65.9 &52.0 &55.0 &69.8 &72.2 \\
SD 3        &66.5 &67.0 &80.3 &76.6 &82.6 &80.8 &81.5 &79.7 &70.1 &75.1 &62.7 &59.2 &62.4 &63.1 &70.1 &69.7 &50.2 &58.9 &65.8 &77.6 &52.0 &28.0 &64.2 &66.4 \\
SD 3.5      &69.5 &68.5 &79.4 &78.9 &79.2 &78.4 &80.7 &84.9 &72.2 &80.6 &67.0 &60.6 &66.1 &63.4 &73.1 &75.8 &59.3 &61.0 &74.6 &62.4 &55.0 &43.0 &66.0 &65.6 \\
SANA Sprint &64.4 &57.9 &75.7 &72.4 &74.6 &70.4 &82.8 &73.4 &71.7 &70.8 &58.5 &51.0 &56.6 &54.1 &69.4 &67.5 &61.4 &64.3 &81.1 &58.9 &8.4 &7.4 &66.3 &59.4 \\
SANA 1.5    &66.9 &67.5 &80.6 &76.3 &81.2 &76.8 &84.9 &84.6 &74.7 &69.1 &62.4 &61.7 &64.3 &55.6 &71.5 &72.0 &64.2 &65.1 &79.3 &81.1 &34.0 &33.0 &72.6 &67.8 \\
PixArt-alpha&45.9 &49.6 &54.6 &62.5 &53.1 &55.8 &65.4 &72.9 &49.8 &53.9 &39.5 &44.2 &37.2 &42.4 &58.4 &51.7 &41.3 &46.5 &51.1 &75.4 &2.0 &0.0 &44.8 &54.3 \\
PixArt-sigma&61.0 &59.7 &69.6 &74.2 &70.8 &77.9 &76.1 &78.4 &66.9 &70.9 &58.7 &48.6 &64.6 &57.6 &66.3 &63.2 &65.6 &55.7 &82.0 &71.3 &1.0 &0.0 &61.2 &53.8 \\
Lumina-Next &52.4 &51.4 &63.7 &67.6 &56.3 &60.4 &68.9 &71.2 &70.4 &66.1 &45.7 &46.8 &50.7 &44.7 &52.3 &58.6 &43.7 &55.9 &69.1 &65.5 &1.0 &0.0 &48.9 &48.1 \\
Hunyuan-DiT &50.4 &54.6 &68.8 &70.6 &65.2 &71.3 &77.1 &75.2 &63.4 &64.6 &43.7 &44.7 &49.3 &42.9 &58.6 &63.4 &47.1 &52.6 &55.4 &72.2 &0.0 &1.0 &39.1 &45.3 \\
\midrule
\addlinespace[-1pt]     
\multicolumn{23}{c}{\textbf{Open-Source \textcolor{middleyellow}{Unified} Models}} \\[-1pt]
\midrule
Show-o   &60.7 &58.2 &72.5 &77.2 &73.7 &81.1 &78.4 &79.4 &66.9 &68.2 &54.6 &49.7 &62.5 &56.3 &69.9 &68.3 &65.7 &57.5 &62.2 &67.6 &7.0 &14.0 &56.4 &49.8 \\
JanusPro &65.5 &66.6 &80.3 &77.1 &78.7 &83.7 &77.9 &74.6 &81.3 &78.4 &60.6 &58.1 &65.1 &57.7 &69.4 &72.4 &68.1 &59.4 &61.3 &69.2 &27.0 &32.0 &67.4 &59.3 \\
Show-o2     &69.2 &67.7 &76.5 &74.7 &72.7 &71.8 &82.6 &81.4 &73.9 &72.9 &65.2 &63.6 &72.4 &71.7 &67.6 &66.2 &70.2 &69.1 &90.0 &89.7 &22.0 &20.0 &66.5 &65.2 \\
Lumina-mGPT &66.1 &64.6 &77.7 &76.1 &72.7 &71.9 &83.4 &82.3 &76.8 &75.4 &61.5 &59.6 &65.5 &64.8 &65.9 &64.9 &64.7 &63.5 &100.0 &99.5 &8.0 &5.0 &56.6 &55.1 \\
Lumina-mGPT2 &58.4 &56.8 &68.5 &67.2 &66.0 &64.8 &75.2 &74.0 &64.3 &63.1 &51.8 &50.5 &61.2 &60.0 &55.1 &53.9 &60.4 &59.2 &73.3 &72.4 &6.0 &5.0 &62.2 &61.0 \\
Bagel        &71.0 &71.3 &81.5 &80.3 &82.8 &83.2 &82.7 &80.2 &79.6 &76.9 &70.0 &72.4 &74.1 &75.2 &67.0 &70.3 &71.8 &75.1 &86.4 &83.6 &24.0 &32.0 &68.0 &68.2 \\
Emu3.5       &85.9 &84.8 &84.8 &83.6 &84.0 &82.9 &89.9 &88.7 &80.5 &79.4 &87.5 &86.3 &83.3 &82.1 &79.7 &78.6 &81.3 &80.1 &100.0 &99.2 &80.0 &80.0 &85.7 &84.5 \\
Emu3.5-Image &88.4 &87.1 &86.2 &85.1 &85.3 &84.2 &90.2 &89.1 &83.2 &82.1 &90.6 &89.4 &88.6 &87.5 &84.5 &83.4 &83.8 &82.7 &100.0 &99.1 &82.0 &83.0 &90.5 &89.4 \\
\midrule
\addlinespace[-1pt]     
\multicolumn{23}{c}{\textbf{\textcolor{attrred}{Closed-Source} Models}} \\[-1pt]
\midrule
DALL-E 3      &72.2 &70.2 &79.7 &77.8 &80.6 &79.1 &82.3 &77.9 &74.8 &78.2 &72.4 &68.5 &74.7 &66.6 &71.1 &72.4 &62.5 &61.7 &88.8 &87.5 &44.0 &52.0 &72.0 &62.5 \\
MidJourney v7 &67.7 &62.7 &76.5 &77.6 &78.9 &81.3 &83.5 &75.8 &73.9 &68.4 &65.6 &59.3 &66.2 &63.9 &82.2 &70.7 &62.1 &63.7 &84.6 &78.7 &42.0 &26.0 &70.3 &62.7 \\
GPT-Image-1   &86.1 &87.6 &89.7 &90.6 &92.4 &86.3 &83.7 &86.0 &95.4 &98.7 &87.5 &89.8 &86.3 &90.2 &86.4 &85.0 &84.7 &84.6 &91.2 &92.2 &79.0 &81.0 &90.8 &92.5 \\
Nano-Banana   &89.2 &88.5 &89.7 &89.3 &84.7 &84.4 &91.0 &91.4 &93.2 &93.0 &90.4 &90.1 &87.7 &87.9 &87.7 &88.0 &88.0 &88.3 &96.7 &96.9 &82.0 &82.0 &92.9 &93.2 \\
NB-Pro        &93.1 &93.3 &93.2 &93.5 &90.7 &91.0 &94.0 &94.3 &94.8 &94.5 &93.6 &93.9 &91.6 &91.9 &91.1 &91.4 &90.9 &91.2 &96.7 &97.0 &84.0 &85.0 &95.2 &95.5 \\
\bottomrule
\end{tabular}
\end{adjustbox}

%% file: tables/genai-bench-8models.tex
\renewcommand{\arraystretch}{1.3}  
\setlength{\tabcolsep}{8pt}       

\centering
\small
\begin{adjustbox}{max width=0.8\linewidth}
\begin{tabular}{
  >{\raggedright\arraybackslash}m{3cm}
  >{\centering\arraybackslash}m{1.2cm}
  >{\centering\arraybackslash}m{1.2cm}
  >{\centering\arraybackslash}m{1.2cm}
}
\toprule
\textbf{Model} & \textbf{Basic} & \textbf{Advanced} & \textbf{Overall} \\
\midrule
SD 3.5\cite{esserScalingRectifiedFlow2024}           & 0.88   & 0.65   & 0.75   \\
SANA 1.5\cite{xieSANA15Efficient2025}         & 0.86   & 0.66   & 0.75   \\
PixArt-Sigma\cite{chenPixArtSWeaktoStrongTraining2024}     & 0.86   & 0.65   & 0.75   \\
Janus-Pro\cite{chenJanusProUnifiedMultimodal}        & 0.80   & 0.64   & 0.71   \\
DALLE-3\cite{betker2023improving}          & 0.85   & 0.76   & 0.81   \\
MidJourney v7\cite{Midjourney}    & 0.85   & 0.68   & 0.75   \\
Nano-Banana\cite{nano-banana}       & 0.84   & 0.74   & 0.79   \\
GPT-Image-1\cite{hurst2024gpt}           & 0.86   & 0.83   & 0.85   \\
\bottomrule
\end{tabular}
\end{adjustbox}

%% file: tables/compbench-8models.tex
\newcommand{\thinrule}{\textcolor{black!60}{\vrule width 0.03pt}}
\newcolumntype{!}{@{\hspace{2pt}\thinrule\hspace{2pt}}}
\renewcommand{\arraystretch}{1.4}
\setlength{\tabcolsep}{2.5pt}

\small
\begin{adjustbox}{width=\linewidth}
\begin{tabular}{
  >{\raggedright\arraybackslash}m{2.8cm}!  
  >{\centering\arraybackslash}m{0.8cm} >{\centering\arraybackslash}m{0.8cm}!  
  >{\centering\arraybackslash}m{0.8cm} >{\centering\arraybackslash}m{0.8cm}!  
  >{\centering\arraybackslash}m{0.8cm} >{\centering\arraybackslash}m{0.8cm}!  
  >{\centering\arraybackslash}m{0.8cm} >{\centering\arraybackslash}m{0.8cm}!  
  >{\centering\arraybackslash}m{0.8cm} >{\centering\arraybackslash}m{0.8cm}!  
  >{\centering\arraybackslash}m{0.8cm} >{\centering\arraybackslash}m{0.8cm}!  
  >{\centering\arraybackslash}m{0.8cm} >{\centering\arraybackslash}m{0.8cm}!  
  >{\centering\arraybackslash}m{0.8cm} >{\centering\arraybackslash}m{0.8cm}!  
  >{\centering\arraybackslash}m{0.8cm} >{\centering\arraybackslash}m{0.8cm}   
}
\toprule
\multirow{3}{*}{\textbf{Model}} &
\multicolumn{2}{c!}{\textbf{AVG}} &
\multicolumn{2}{c!}{\textbf{Color}} &
\multicolumn{2}{c!}{\textbf{Shape}} &
\multicolumn{2}{c!}{\textbf{Texture}} &
\multicolumn{2}{c!}{\textbf{Numeracy}} &
\multicolumn{2}{c!}{\textbf{2D Spatial}} &
\multicolumn{2}{c!}{\textbf{3D Spatial}} &
\multicolumn{2}{c!}{\textbf{Non-Spatial}} &
\multicolumn{2}{c}{\textbf{Complex}} \\
\cmidrule(lr{.3em}){2-3}
\cmidrule(lr{.3em}){4-5}
\cmidrule(lr{.3em}){6-7}
\cmidrule(lr{.3em}){8-9}
\cmidrule(lr{.3em}){10-11}
\cmidrule(lr{.3em}){12-13}
\cmidrule(lr{.3em}){14-15}
\cmidrule(lr{.3em}){16-17}
\cmidrule(l){18-19}
& Expert & GPT & Expert & GPT & Expert & GPT & Expert & GPT & Expert & GPT & Expert & GPT & Expert & GPT & Expert & GPT & Expert & GPT \\
\midrule
SD 3.5\cite{esserScalingRectifiedFlow2024}           &0.507 &68.89 & 0.767 & 83.40 & 0.596 & 85.15 & 0.706 & 87.44 & 0.621 & 45.03 & 0.277 & 44.76 & 0.403 & 45.06 & 0.315 & 85.03 & 0.376 & 75.26 \\
SANA 1.5\cite{xieSANA15Efficient2025}         &0.507 &69.72 & 0.755 & 88.83 & 0.536 & 79.59 & 0.686 & 78.92 & 0.611 & 53.10 & 0.362 & 51.03 & 0.412 & 44.26 & 0.312 & 84.46 & 0.382 & 77.56 \\
PixArt-Sigma\cite{chenPixArtSWeaktoStrongTraining2024}     &0.436 &56.92 & 0.587 & 82.60 & 0.476 & 62.45 & 0.569 & 71.08 & 0.549 & 30.10 & 0.247 & 46.83 & 0.366 & 27.00 & 0.308 & 67.40 & 0.383 & 67.90 \\
Janus-Pro\cite{chenJanusProUnifiedMultimodal}        &0.285 & 47.34 &  0.393 & 57.45 & 0.264 & 42.34 & 0.349 & 48.49 & 0.335 & 36.90  & 0.080 & 17.86 & 0.209 & 29.83 & 0.299 & 71.36 & 0.355 & 74.56 \\
DALLE-3\cite{betker2023improving}          &0.511 &76.94 & 0.803 & 84.03 & 0.637 & 85.15 & 0.774 & 86.88 & 0.617 & 62.90 & 0.253 & 63.20 & 0.353 & 68.06 & 0.297 & 85.73 & 0.353 & 79.58 \\
MidJourney v7\cite{Midjourney}  &0.507 &70.93 & 0.750 & 84.33 & 0.525 & 72.68 & 0.787 & 85.22 & 0.670 & 55.22 & 0.278 & 52.27 & 0.384 & 51.06 & 0.310 & 89.80 & 0.349 & 76.88 \\
Nano-Banana\cite{nano-banana}       &0.550 &87.76 & 0.760 & 92.20 & 0.611 & 81.39 & 0.790 & 93.87 & 0.779 & 76.92 & 0.433 & 84.66 & 0.393 & 67.23 & 0.316 & 89.84 & 0.379 & 92.11 \\
GPT-Image-1\cite{ye2025echo4oharnessingpowergpt4o}       &0.572 &86.59 & 0.795 & 95.34 & 0.593 & 84.30 & 0.831 & 91.04 & 0.800 & 79.86 & 0.450 & 85.83 & 0.406 & 70.46 & 0.311 & 95.50 & 0.389 & 90.40 \\
\bottomrule
\end{tabular}
\end{adjustbox}

%% file: tables/compbenchpp_Spearman.tex
\newcommand{\thinrule}{\textcolor{black!50}{\vrule width 0.03pt}}
\newcolumntype{!}{@{\hspace{4pt}\thinrule\hspace{4pt}}}
\renewcommand{\arraystretch}{1.5}
\setlength{\tabcolsep}{4pt}

\small
\centering
\vspace{2pt}
\begin{adjustbox}{width=\linewidth}
\begin{tabular}{
  >{\raggedright\arraybackslash}m{1.6cm}!  
  >{\centering\arraybackslash}m{1.5cm}!    
  >{\centering\arraybackslash}m{1.5cm}!    
  >{\centering\arraybackslash}m{1.5cm}!    
  >{\centering\arraybackslash}m{1.5cm}!    
  >{\centering\arraybackslash}m{1.5cm}!    
  >{\centering\arraybackslash}m{1.5cm}!    
  >{\centering\arraybackslash}m{1.5cm}!    
  >{\centering\arraybackslash}m{1.5cm}     
}
\toprule
\textbf{Comp++} & \textbf{Color} & \textbf{Shape} & \textbf{Texture} & \textbf{Numeracy} & \textbf{2D Spatial} & \textbf{3D Spatial} & \textbf{Non-Spatial} & \textbf{Complex} \\
\midrule
\multicolumn{9}{c}{\textit{Spearman $\rho$}}\\[-1pt]
\midrule
\textbf{Experts} & 0.58 & 0.78 & 0.61 & 0.24 & 0.00 & 0.14 & 0.07 & 0.36 \\
\textbf{GPT}    & 0.36 & 0.49 & 0.43 & 0.60 & 0.60 & 0.79 & 0.60 & 0.52 \\
\bottomrule
\end{tabular}
\end{adjustbox}

%% file: tables/TIIF-Bench-Spearman.tex
\newcommand{\thinrule}{\textcolor{black!60}{\vrule width 0.03pt}}
\newcolumntype{!}{@{\hspace{4pt}\thinrule\hspace{4pt}}}
\renewcommand{\arraystretch}{1.7} 
\setlength{\tabcolsep}{3pt}

\small
\centering
\begin{adjustbox}{width=\linewidth}
\begin{tabular}{
  >{\raggedright\arraybackslash}m{2.3cm}!       
  *{6}{>{\centering\arraybackslash}m{0.7cm}!}   
  *{10}{>{\centering\arraybackslash}m{0.7cm}!}  
  >{\centering\arraybackslash}m{0.7cm}!         
  >{\centering\arraybackslash}m{0.7cm}          
}
\toprule
\multirow{3}{*}{\textbf{Model}} &
\multicolumn{6}{c!}{\textbf{Basic Following}} &
\multicolumn{10}{c!}{\textbf{Advanced Following}} &
\multicolumn{2}{c}{\textbf{Real-World}} \\ \cmidrule(lr{.3em}){2-7}\cmidrule(lr{.3em}){8-17}\cmidrule(l){18-19}
& \multicolumn{2}{c!}{Attribute} & \multicolumn{2}{c!}{Relation} & \multicolumn{2}{c!}{Reasoning}%
& \multicolumn{2}{c!}{\makecell{Attribute\\+ Relation}} & \multicolumn{2}{c!}{\makecell{Attribute\\+ Reasoning}} & \multicolumn{2}{c!}{\makecell{Relation\\+ Reasoning}}%
& \multicolumn{2}{c!}{Style} & \multicolumn{2}{c!}{Text}%
& \multicolumn{2}{c}{Complex} \\ \cmidrule(lr{.3em}){2-7}\cmidrule(lr{.3em}){8-17}\cmidrule(l){18-19}
& short & long & short & long & short & long & short & long & short & long & short & long & short & long & short & long & short & long \\ 
\midrule
\addlinespace[-1pt]     
\multicolumn{19}{c}{\textit{Spearman $\rho$}} \\
\midrule

VLM Eval/PNED   &0.81 &0.88 &0.88 &0.91 &0.93 &1.00 &0.93 &0.95 &0.93 &0.98 &0.95 &1.00 &0.81 &0.81 &0.98 &1.00 &0.85 &0.81 \\
\bottomrule
\end{tabular}
\end{adjustbox}

%% file: tables/attr_pools_table.tex
\small
\setlength{\tabcolsep}{5pt}
\renewcommand{\arraystretch}{1.8}
\newcommand{\thinrule}{\textcolor{black!60}{\vrule width 0.01pt}}
\newcolumntype{!}{@{\hspace{6pt}\thinrule\hspace{2pt}}}

\begin{adjustbox}{width=\linewidth}
\begin{tabular}{
  >{\raggedright\arraybackslash}m{2cm}!   
  >{\raggedright\arraybackslash}m{2.2cm}! 
  >{\raggedright\arraybackslash}m{7.0cm}! 
  >{\raggedright\arraybackslash}m{4.8cm}  
}
\toprule
\multicolumn{1}{>{\centering\arraybackslash}m{2cm}!}{\textbf{Category}} &
\multicolumn{1}{>{\centering\arraybackslash}m{2.2cm}!}{\textbf{Dimension}} &
\multicolumn{1}{>{\centering\arraybackslash}m{7.0cm}!}{\textbf{Example Prompt}} &
\multicolumn{1}{>{\centering\arraybackslash}m{4.8cm}}{\textbf{Extracted Attributes}} \\

\midrule
\multirow[c]{3}{*}[-3.8ex]{\parbox[c]{2cm}{\centering \textbf{Attributes}}}
 & Color      & A tan dog with sky blue eyes posing for a picture with a man sitting on a chair in the background. & \textcolor{attrred}{tan} \textcolor{objblue}{dog}, \textcolor{attrred}{sky blue} \textcolor{objblue}{eyes} \\
 & Shape      & The conical salt and pepper shakers with their cylindrical bases and spherical tops seasoned food in the square dining room. & \textcolor{attrred}{conical} \textcolor{objblue}{salt and pepper shakers}, \textcolor{attrred}{cylindrical} \textcolor{objblue}{bases}, \textcolor{attrred}{spherical} \textcolor{objblue}{tops}, \textcolor{attrred}{square} \textcolor{objblue}{dining room} \\
 & Texture    & The wooden table is covered with a fabric tablecloth and adorned with a glass vase. & \textcolor{attrred}{wooden} \textcolor{objblue}{table}, \textcolor{attrred}{fabric} \textcolor{objblue}{tablecloth}, \textcolor{attrred}{glass} \textcolor{objblue}{vase} \\
\midrule
\multirow[c]{3}{*}{\parbox[c]{2cm}{\centering \textbf{Relations}}}
 & 2D Spatial     & A butterfly on the top of a desk. & \textcolor{objblue}{butterfly} \textcolor{attrred}{on top of} \textcolor{objblue}{desk} \\
 & 3D Perspective & A key in front of a girl. & \textcolor{objblue}{key} \textcolor{attrred}{in front of} \textcolor{objblue}{girl} \\
 & Action         & A dragon perched majestically on a craggy, smoke-wreathed mountain. & \textcolor{objblue}{dragon perched on mountain} \\
\midrule
\multirow[c]{4}{*}[-7.8ex]{\parbox[c]{2cm}{\centering \textbf{Reasoning}}}
 & Numeracy        & Three bottles stood next to three printers on the shelf. & \textcolor{attrred}{three} \textcolor{objblue}{bottles}, \textcolor{attrred}{three} \textcolor{objblue}{printers} \\
 & Differentiation & A person in uniform pointing out landmarks to a person in a window seat wearing noise-canceling headphones. & \textcolor{objblue}{a person in uniform, another person wearing headphones} \\
 & Comparison      & In a mysterious swamp, the flowers are taller than the trees. & \textcolor{objblue}{flowers are taller than the trees} \\
 & Negation        & The girl with glasses is drawing, and the girl without glasses is singing. & \textcolor{objblue}{girl without glasses is singing} \\
\bottomrule
\end{tabular}
\end{adjustbox}

%% file: tables/stats.tex

\newcommand{\thinrule}{\textcolor{black!60}{\vrule width 0.03pt}}
\newcolumntype{!}{@{\hspace{4pt}\thinrule\hspace{4pt}}}
\renewcommand{\arraystretch}{1.7} 
\setlength{\tabcolsep}{3pt}

\small
\centering
\begin{adjustbox}{width=\linewidth}
\begin{tabular}{
  >{\raggedright\arraybackslash}m{2.0cm}!       
  *{6}{>{\centering\arraybackslash}m{0.7cm}!}   
  *{10}{>{\centering\arraybackslash}m{0.7cm}!}  
  >{\centering\arraybackslash}m{0.7cm}!         
  >{\centering\arraybackslash}m{0.7cm}          
}
\toprule
\multirow{3}{*}{\textbf{Statistic}} &
\multicolumn{6}{c!}{\textbf{Basic Following}} &
\multicolumn{10}{c!}{\textbf{Advanced Following}} &
\multicolumn{2}{c}{\textbf{Real-World}} \\ \cmidrule(lr{.3em}){2-7}\cmidrule(lr{.3em}){8-17}\cmidrule(l){18-19}
& \multicolumn{2}{c!}{Attribute} & \multicolumn{2}{c!}{Relation} & \multicolumn{2}{c!}{Reasoning}%
& \multicolumn{2}{c!}{\makecell{Attribute\\+ Relation}} & \multicolumn{2}{c!}{\makecell{Attribute\\+ Reasoning}} & \multicolumn{2}{c!}{\makecell{Relation\\+ Reasoning}}%
& \multicolumn{2}{c!}{Style} & \multicolumn{2}{c!}{Text}%
& \multicolumn{2}{c}{Complex} \\ \cmidrule(lr{.3em}){2-7}\cmidrule(lr{.3em}){8-17}\cmidrule(l){18-19}
& short & long & short & long & short & long & short & long & short & long & short & long & short & long & short & long & short & long \\ 
\midrule
Avg. Len.
& 14.7 & 58.8 & 9.6 & 51.8 & 14.2 & 64.2
& 10.4 & 55.2 & 15.0 & 66.6 & 15.4 & 61.9
& 10.3 & 131.7 & 25.1 & 191.4
& 66.3 & 156.1 \\
\addlinespace[2pt]
Check Len.
& \multicolumn{2}{c!}{5.0} & \multicolumn{2}{c!}{3.9} & \multicolumn{2}{c!}{4.1}
& \multicolumn{2}{c!}{5.1} & \multicolumn{2}{c!}{6.1} & \multicolumn{2}{c!}{5.0}
& \multicolumn{2}{c!}{--} & \multicolumn{2}{c!}{--}
& \multicolumn{2}{c}{8.1} \\
\addlinespace[-1pt]

\bottomrule
\end{tabular}
\end{adjustbox}

%% file: tables/text_gen_results.tex
\newcommand{\thinrule}{\textcolor{black!50}{\vrule width 0.03pt}}
\newcolumntype{!}{@{\hspace{4pt}\thinrule\hspace{4pt}}}
\renewcommand{\arraystretch}{1.2}
\setlength{\tabcolsep}{4pt}

\small
\centering
\begin{adjustbox}{max width=1.0\linewidth, center}
\begin{tabular}{
  >{\raggedright\arraybackslash}m{3.0cm}!    
  >{\centering\arraybackslash}m{1.8cm}        
  >{\centering\arraybackslash}m{1.8cm}!       
  >{\centering\arraybackslash}m{1.8cm}        
  >{\centering\arraybackslash}m{1.8cm}        
}
\toprule
\multirow{2}{*}{\textbf{Model}} &
\multicolumn{2}{c!}{\textbf{GNED}~\ensuremath{\downarrow}} &
\multicolumn{2}{c}{\textbf{Recall}~\ensuremath{\uparrow}} \\
\cmidrule(lr{.3em}){2-3} \cmidrule(l){4-5}
& \textbf{Short} & \textbf{Long} & \textbf{Short} & \textbf{Long}\\
\midrule
\multicolumn{5}{c}{\textbf{\textcolor{objblue}{Diffusion based Open-Source Models}}} \\[-1pt]
\midrule
SD XL  \cite{podellSDXLImprovingLatent2023a}  &0.85   &0.97  &12.56  &1.00   \\
SD 3 \cite{esserScalingRectifiedFlow2024}  &0.48   &0.72  &57.17  &24.78  \\
SD 3.5  \cite{esserScalingRectifiedFlow2024}  &0.45   &0.57  &67.71  &46.19  \\
SANA 1.5  \cite{xieSANA15Efficient2025}  &0.66   &0.67  &19.51  &16.91  \\
Infinity  \cite{Infinity}   &0.77   &0.78  &14.87  &11.90  \\
Flux.1 Dev \cite{flux2024}    &0.48   &0.45  &53.54  &58.65  \\

\midrule
\multicolumn{5}{c}{\textbf{\textcolor{middleyellow}{Open-Source Unified Models}}} \\[-1pt]
\midrule
Show-o   \cite{xieShowoOneSingle2024}       &0.93  &0.86  &0.00  &0.00   \\
JanusPro \cite{chenJanusProUnifiedMultimodal}       &0.73  &0.68  &15.42  &19.06   \\

\midrule
\multicolumn{5}{c}{\textbf{\textcolor{attrred}{Closed-Source Models}}} \\[-1pt]
\midrule
DALLE 3 \cite{betker2023improving}    &0.56   &0.48 &53.02  &61.22  \\
MidJourney v7 \cite{Midjourney}  &0.68   &0.74  &20.23  &9.22      \\
Nano-Banana\cite{nano-banana}   &0.21   &0.19  &82.55  &80.84  \\
GPT-Image-1 \cite{hurst2024gpt}  &0.18   &0.18  &84.58  &82.22 \\
\bottomrule
\end{tabular}
\end{adjustbox}

%% file: main.bbl
\begin{thebibliography}{10}
\providecommand{\url}[1]{\texttt{#1}}
\providecommand{\urlprefix}{URL }
\providecommand{\doi}[1]{https://doi.org/#1}

\bibitem{an2026genius}
An, R., Yang, S., Guo, Z., Dai, W., Shen, Z., Li, H., Zhang, R., Wei, X., Li, G., Wu, W., Zhang, W.: {GENIUS}: Generative fluid intelligence evaluation suite (2026), \url{https://arxiv.org/abs/2602.11144}

\bibitem{bai2025qwen3vltechnicalreport}
Bai, S., Cai, Y., Chen, R., Chen, K., Chen, X., Cheng, Z., Deng, L., Ding, W., Gao, C., Ge, C., Ge, W., Guo, Z., Huang, Q., Huang, J., Huang, F., Hui, B., Jiang, S., Li, Z., Li, M., Li, M., Li, K., Lin, Z., Lin, J., Liu, X., Liu, J., Liu, C., Liu, Y., Liu, D., Liu, S., Lu, D., Luo, R., Lv, C., Men, R., Meng, L., Ren, X., Ren, X., Song, S., Sun, Y., Tang, J., Tu, J., Wan, J., Wang, P., Wang, P., Wang, Q., Wang, Y., Xie, T., Xu, Y., Xu, H., Xu, J., Yang, Z., Yang, M., Yang, J., Yang, A., Yu, B., Zhang, F., Zhang, H., Zhang, X., Zheng, B., Zhong, H., Zhou, J., Zhou, F., Zhou, J., Zhu, Y., Zhu, K.: Qwen3-vl technical report (2025), \url{https://arxiv.org/abs/2511.21631}

\bibitem{baldridge2024imagen}
Baldridge, J., Bauer, J., Bhutani, M., Brichtova, N., Bunner, A., Castrejon, L., Chan, K., Chen, Y., Dieleman, S., Du, Y., et~al.: Imagen 3. arXiv preprint arXiv:2408.07009  (2024)

\bibitem{betkerImprovingImageGeneration}
Betker, J., Goh, G., Jing, L., Brooks, T., Wang, J., Li, L., Ouyang, L., Zhuang, J., Lee, J., Guo, Y., Manassra, W., Dhariwal, P., Chu, C., Jiao, Y., Ramesh, A.: Improving {{Image Generation}} with {{Better Captions}}

\bibitem{betker2023improving}
Betker, J., Goh, G., Jing, L., Brooks, T., Wang, J., Li, L., Ouyang, L., Zhuang, J., Lee, J., Guo, Y., et~al.: Improving image generation with better captions. Computer Science. https://cdn. openai. com/papers/dall-e-3. pdf  \textbf{2}(3), ~8 (2023)

\bibitem{chang2025oneigbenchomnidimensionalnuancedevaluation}
Chang, J., Fang, Y., Xing, P., Wu, S., Cheng, W., Wang, R., Zeng, X., Yu, G., Chen, H.B.: Oneig-bench: Omni-dimensional nuanced evaluation for image generation (2025), \url{https://arxiv.org/abs/2506.07977}

\bibitem{chenPixArtSWeaktoStrongTraining2024}
Chen, J., Ge, C., Xie, E., Wu, Y., Yao, L., Ren, X., Wang, Z., Luo, P., Lu, H., Li, Z.: {{PixArt-$\Sigma$}}: {{Weak-to-Strong Training}} of {{Diffusion Transformer}} for {{4K Text-to-Image Generation}} (Mar 2024). \doi{10.48550/arXiv.2403.04692}

\bibitem{chenPIXARTdFastControllable2024}
Chen, J., Wu, Y., Luo, S., Xie, E., Paul, S., Luo, P., Zhao, H., Li, Z.: {{PIXART-$\delta$}}: {{Fast}} and {{Controllable Image Generation}} with {{Latent Consistency Models}} (Jan 2024). \doi{10.48550/arXiv.2401.05252}

\bibitem{chenSANASprintOneStepDiffusion2025}
Chen, J., Xue, S., Zhao, Y., Yu, J., Paul, S., Chen, J., Cai, H., Xie, E., Han, S.: {{SANA-Sprint}}: {{One-Step Diffusion}} with {{Continuous-Time Consistency Distillation}} (Mar 2025). \doi{10.48550/arXiv.2503.09641}

\bibitem{chenPixArt$a$FastTraining2023}
Chen, J., Yu, J., Ge, C., Yao, L., Xie, E., Wu, Y., Wang, Z., Kwok, J., Luo, P., Lu, H., Li, Z.: {{PixArt-}}\${$\alpha\$$}: {{Fast Training}} of {{Diffusion Transformer}} for {{Photorealistic Text-to-Image Synthesis}} (Dec 2023). \doi{10.48550/arXiv.2310.00426}

\bibitem{chen2025r2ibenchbenchmarkingreasoningdriventexttoimage}
Chen, K., Lin, Z., Xu, Z., Shen, Y., Yao, Y., Rimchala, J., Zhang, J., Huang, L.: R2i-bench: Benchmarking reasoning-driven text-to-image generation (2025), \url{https://arxiv.org/abs/2505.23493}

\bibitem{chenJanusProUnifiedMultimodal}
Chen, X., Wu, Z., Liu, X., Pan, Z., Liu, W., Xie, Z., Yu, X., Ruan, C.: Janus-{{Pro}}: {{Unified Multimodal Understanding}} and {{Generation}} with {{Data}} and {{Model Scaling}}

\bibitem{cho2024davidsonianscenegraphimproving}
Cho, J., Hu, Y., Garg, R., Anderson, P., Krishna, R., Baldridge, J., Bansal, M., Pont-Tuset, J., Wang, S.: Davidsonian scene graph: Improving reliability in fine-grained evaluation for text-to-image generation (2024), \url{https://arxiv.org/abs/2310.18235}

\bibitem{Cui_2024_CVPR}
Cui, S., Guo, J., An, X., Deng, J., Zhao, Y., Wei, X., Feng, Z.: {IDAdapter}: Learning mixed features for tuning-free personalization of text-to-image models. In: Proceedings of the IEEE/CVF Conference on Computer Vision and Pattern Recognition (CVPR) Workshops. pp. 950--959 (June 2024)

\bibitem{cui2025emu35nativemultimodalmodels}
Cui, Y., Chen, H., Deng, H., Huang, X., Li, X., Liu, J., Liu, Y., Luo, Z., Wang, J., Wang, W., Wang, Y., Wang, C., Zhang, F., Zhao, Y., Pan, T., Li, X., Hao, Z., Ma, W., Chen, Z., Ao, Y., Huang, T., Wang, Z., Wang, X.: Emu3.5: Native multimodal models are world learners (2025), \url{https://arxiv.org/abs/2510.26583}

\bibitem{deng2025emergingpropertiesunifiedmultimodal}
Deng, C., Zhu, D., Li, K., Gou, C., Li, F., Wang, Z., Zhong, S., Yu, W., Nie, X., Song, Z., Shi, G., Fan, H.: Emerging properties in unified multimodal pretraining (2025), \url{https://arxiv.org/abs/2505.14683}

\bibitem{esserScalingRectifiedFlow2024}
Esser, P., Kulal, S., Blattmann, A., Entezari, R., M{\"u}ller, J., Saini, H., Levi, Y., Lorenz, D., Sauer, A., Boesel, F., Podell, D., Dockhorn, T., English, Z., Rombach, R.: Scaling {{Rectified Flow Transformers}} for {{High-Resolution Image Synthesis}}. In: Forty-First {{International Conference}} on {{Machine Learning}} (Jun 2024)

\bibitem{ghosh2023genevalobjectfocusedframeworkevaluating}
Ghosh, D., Hajishirzi, H., Schmidt, L.: Geneval: An object-focused framework for evaluating text-to-image alignment (2023), \url{https://arxiv.org/abs/2310.11513}

\bibitem{nano-banana}
Google: nano-banana. \url{https://developers.googleblog.com/introducing-gemini-2-5-flash-image/} (2025)

\bibitem{nano-banana-pro}
Google: nano-banana-pro. \url{https://deepmind.google/models/gemini-image/pro/} (2025)

\bibitem{nano-banana-2}
Google: nano-banana-2. \url{https://deepmind.google/models/gemini-image/flash/} (2026)

\bibitem{Infinity}
Han, J., Liu, J., Jiang, Y., Yan, B., Zhang, Y., Yuan, Z., Peng, B., Liu, X.: Infinity: Scaling bitwise autoregressive modeling for high-resolution image synthesis (2024)

\bibitem{hesselCLIPScoreReferencefreeEvaluation2022}
Hessel, J., Holtzman, A., Forbes, M., Bras, R.L., Choi, Y.: {{CLIPScore}}: {{A Reference-free Evaluation Metric}} for {{Image Captioning}} (Mar 2022). \doi{10.48550/arXiv.2104.08718}

\bibitem{hu2024ellaequipdiffusionmodels}
Hu, X., Wang, R., Fang, Y., Fu, B., Cheng, P., Yu, G.: Ella: Equip diffusion models with llm for enhanced semantic alignment (2024), \url{https://arxiv.org/abs/2403.05135}

\bibitem{hu2023tifaaccurateinterpretabletexttoimage}
Hu, Y., Liu, B., Kasai, J., Wang, Y., Ostendorf, M., Krishna, R., Smith, N.A.: Tifa: Accurate and interpretable text-to-image faithfulness evaluation with question answering (2023), \url{https://arxiv.org/abs/2303.11897}

\bibitem{huang2025t2icompbenchenhancedcomprehensivebenchmark}
Huang, K., Duan, C., Sun, K., Xie, E., Li, Z., Liu, X.: T2i-compbench++: An enhanced and comprehensive benchmark for compositional text-to-image generation (2025), \url{https://arxiv.org/abs/2307.06350}

\bibitem{hurst2024gpt}
Hurst, A., Lerer, A., Goucher, A.P., Perelman, A., Ramesh, A., Clark, A., Ostrow, A., Welihinda, A., Hayes, A., Radford, A., et~al.: Gpt-4o system card. arXiv preprint arXiv:2410.21276  (2024)

\bibitem{imagen-team-googleImagen32024}
{Imagen-Team-Google}, Baldridge, J., Bauer, J., Bhutani, M., Others: Imagen 3 (Dec 2024). \doi{10.48550/arXiv.2408.07009}

\bibitem{kamath2025geneval2addressingbenchmark}
Kamath, A., Chang, K.W., Krishna, R., Zettlemoyer, L., Hu, Y., Ghazvininejad, M.: Geneval 2: Addressing benchmark drift in text-to-image evaluation (2025), \url{https://arxiv.org/abs/2512.16853}

\bibitem{flux2024}
Labs, B.F.: Flux. \url{https://github.com/black-forest-labs/flux} (2024)

\bibitem{flux.2}
Labs, B.F.: Flux.2. \url{https://bfl.ai/models/flux-2} (2026)

\bibitem{li2024genaibenchevaluatingimprovingcompositional}
Li, B., Lin, Z., Pathak, D., Li, J., Fei, Y., Wu, K., Ling, T., Xia, X., Zhang, P., Neubig, G., Ramanan, D.: Genai-bench: Evaluating and improving compositional text-to-visual generation (2024), \url{https://arxiv.org/abs/2406.13743}

\bibitem{li2022mplugeffectiveefficientvisionlanguage}
Li, C., Xu, H., Tian, J., Wang, W., Yan, M., Bi, B., Ye, J., Chen, H., Xu, G., Cao, Z., Zhang, J., Huang, S., Huang, F., Zhou, J., Si, L.: mplug: Effective and efficient vision-language learning by cross-modal skip-connections (2022), \url{https://arxiv.org/abs/2205.12005}

\bibitem{li2022blipbootstrappinglanguageimagepretraining}
Li, J., Li, D., Xiong, C., Hoi, S.: Blip: Bootstrapping language-image pre-training for unified vision-language understanding and generation (2022), \url{https://arxiv.org/abs/2201.12086}

\bibitem{li2025easierpaintingthinkingtexttoimage}
Li, O., Wang, Y., Hu, X., Huang, H., Chen, R., Ou, J., Tao, X., Wan, P., Qi, X., Feng, F.: Easier painting than thinking: Can text-to-image models set the stage, but not direct the play? (2025), \url{https://arxiv.org/abs/2509.03516}

\bibitem{li2026easierpaintingthinkingtexttoimage}
Li, O., Wang, Y., Hu, X., Huang, H., Chen, R., Ou, J., Tao, X., Wan, P., Qi, X., Feng, F.: Easier painting than thinking: Can text-to-image models set the stage, but not direct the play? (2026), \url{https://arxiv.org/abs/2509.03516}

\bibitem{liHunyuanDiTPowerfulMultiResolution2024}
Li, Z., Zhang, J., Lin, Q., Xiong, J., Long, Y., Deng, X., Zhang, Y., Liu, X., Huang, M., Xiao, Z., Chen, D., He, J., Li, J., Li, W., Zhang, C., Quan, R., Lu, J., Huang, J., Yuan, X., Zheng, X., Li, Y., Zhang, J., Zhang, C., Chen, M., Liu, J., Fang, Z., Wang, W., Xue, J., Tao, Y., Zhu, J., Liu, K., Lin, S., Sun, Y., Li, Y., Wang, D., Chen, M., Hu, Z., Xiao, X., Chen, Y., Liu, Y., Liu, W., Wang, D., Yang, Y., Jiang, J., Lu, Q.: Hunyuan-{{DiT}}: {{A Powerful Multi-Resolution Diffusion Transformer}} with {{Fine-Grained Chinese Understanding}} (May 2024). \doi{10.48550/arXiv.2405.08748}

\bibitem{lin2025draw}
Lin, W., Wei, X., An, R., Gao, P., Zou, B., Luo, Y., Huang, S., Zhang, S., Li, H.: Draw-and-understand: Leveraging visual prompts to enable {MLLMs} to comprehend what you want. In: International Conference on Learning Representations (2025)

\bibitem{lin2026perceive}
Lin, W., Wei, X., An, R., Ren, T., Chen, T., Zhang, R., Guo, Z., Zhang, W., Zhang, L., Li, H.: Perceive anything: Recognize, explain, caption, and segment anything in images and videos (2025), \url{https://arxiv.org/abs/2506.05302}

\bibitem{lin2025pixwizard}
Lin, W., Wei, X., Zhang, R., Zhuo, L., Zhao, S., Huang, S., Xie, J., Gao, P., Li, H.: {PixWizard}: Versatile image-to-image visual assistant with open-language instructions. In: International Conference on Learning Representations (2025)

\bibitem{lin2024evaluatingtexttovisualgenerationimagetotext}
Lin, Z., Pathak, D., Li, B., Li, J., Xia, X., Neubig, G., Zhang, P., Ramanan, D.: Evaluating text-to-visual generation with image-to-text generation (2024), \url{https://arxiv.org/abs/2404.01291}

\bibitem{liu2025_Luminamgpt_illuminateflexiblephotorealistic}
Liu, D., Zhao, S., Zhuo, L., Lin, W., Xin, Y., Li, X., Qin, Q., Qiao, Y., Li, H., Gao, P.: Lumina-mgpt: Illuminate flexible photorealistic text-to-image generation with multimodal generative pretraining (2025), \url{https://arxiv.org/abs/2408.02657}

\bibitem{ma2025hpsv3widespectrumhumanpreference}
Ma, Y., Shui, Y., Wu, X., Sun, K., Li, H.: Hpsv3: Towards wide-spectrum human preference score (2025), \url{https://arxiv.org/abs/2508.03789}

\bibitem{niu2025wiseworldknowledgeinformedsemantic}
Niu, Y., Ning, M., Zheng, M., Jin, W., Lin, B., Jin, P., Liao, J., Feng, C., Ning, K., Zhu, B., Yuan, L.: Wise: A world knowledge-informed semantic evaluation for text-to-image generation (2025), \url{https://arxiv.org/abs/2503.07265}

\bibitem{podellSDXLImprovingLatent2023a}
Podell, D., English, Z., Lacey, K., Blattmann, A., Dockhorn, T., M{\"u}ller, J., Penna, J., Rombach, R.: {{SDXL}}: {{Improving Latent Diffusion Models}} for {{High-Resolution Image Synthesis}} (Jul 2023). \doi{10.48550/arXiv.2307.01952}

\bibitem{qin2025luminaimage20unifiedefficient}
Qin, Q., Zhuo, L., Xin, Y., Du, R., Li, Z., Fu, B., Lu, Y., Yuan, J., Li, X., Liu, D., Zhu, X., Zhang, M., Beddow, W., Millon, E., Perez, V., Wang, W., He, C., Zhang, B., Liu, X., Li, H., Qiao, Y., Xu, C., Gao, P.: Lumina-image 2.0: A unified and efficient image generative framework (2025), \url{https://arxiv.org/abs/2503.21758}

\bibitem{qwen2025qwen25technicalreport}
Qwen, :, Yang, A., Yang, B., Zhang, B., Hui, B., Zheng, B., Yu, B., Li, C., Liu, D., Huang, F., Wei, H., Lin, H., Yang, J., Tu, J., Zhang, J., Yang, J., Yang, J., Zhou, J., Lin, J., Dang, K., Lu, K., Bao, K., Yang, K., Yu, L., Li, M., Xue, M., Zhang, P., Zhu, Q., Men, R., Lin, R., Li, T., Tang, T., Xia, T., Ren, X., Ren, X., Fan, Y., Su, Y., Zhang, Y., Wan, Y., Liu, Y., Cui, Z., Zhang, Z., Qiu, Z.: Qwen2.5 technical report (2025), \url{https://arxiv.org/abs/2412.15115}

\bibitem{radford2021learningtransferablevisualmodels}
Radford, A., Kim, J.W., Hallacy, C., Ramesh, A., Goh, G., Agarwal, S., Sastry, G., Askell, A., Mishkin, P., Clark, J., Krueger, G., Sutskever, I.: Learning transferable visual models from natural language supervision (2021), \url{https://arxiv.org/abs/2103.00020}

\bibitem{rombachHighResolutionImageSynthesis2022}
Rombach, R., Blattmann, A., Lorenz, D., Esser, P., Ommer, B.: High-{{Resolution Image Synthesis With Latent Diffusion Models}}. In: Proceedings of the {{IEEE}}/{{CVF Conference}} on {{Computer Vision}} and {{Pattern Recognition}}. pp. 10684--10695 (2022)

\bibitem{dinov3}
Siméoni, O., Vo, H.V., Seitzer, M., Baldassarre, F., Oquab, M., Jose, C., Khalidov, V., Szafraniec, M., Yi, S., Ramamonjisoa, M., Massa, F., Haziza, D., Wehrstedt, L., Wang, J., Darcet, T., Moutakanni, T., Sentana, L., Roberts, C., Vedaldi, A., Tolan, J., Brandt, J., Couprie, C., Mairal, J., Jégou, H., Labatut, P., Bojanowski, P.: Dinov3 (2025), \url{https://arxiv.org/abs/2508.10104}

\bibitem{somepalli2024measuringstylesimilaritydiffusion}
Somepalli, G., Gupta, A., Gupta, K., Palta, S., Goldblum, M., Geiping, J., Shrivastava, A., Goldstein, T.: Measuring style similarity in diffusion models (2024), \url{https://arxiv.org/abs/2404.01292}

\bibitem{sun2025t2ireasonbenchbenchmarkingreasoninginformedtexttoimage}
Sun, K., Fang, R., Duan, C., Liu, X., Liu, X.: T2i-reasonbench: Benchmarking reasoning-informed text-to-image generation (2025), \url{https://arxiv.org/abs/2508.17472}

\bibitem{sunAutoregressiveModelBeats2024}
Sun, P., Jiang, Y., Chen, S., Zhang, S., Peng, B., Luo, P., Yuan, Z.: Autoregressive {{Model Beats Diffusion}}: {{Llama}} for {{Scalable Image Generation}} (Jun 2024). \doi{10.48550/arXiv.2406.06525}

\bibitem{Midjourney}
Team, M.: Midjourney. \url{https://www.midjourney.com/} (2025)

\bibitem{VAR}
Tian, K., Jiang, Y., Yuan, Z., Peng, B., Wang, L.: Visual autoregressive modeling: Scalable image generation via next-scale prediction (2024)

\bibitem{tong2026delving}
Tong, C., Guo, Z., Zhang, R., Shan, W., Wei, X., Xing, Z., Li, H., Heng, P.A.: Delving into {RL} for image generation with {CoT}: A study on {DPO} vs. {GRPO}. In: Advances in Neural Information Processing Systems (2025)

\bibitem{tschannen2025siglip2multilingualvisionlanguage}
Tschannen, M., Gritsenko, A., Wang, X., Naeem, M.F., Alabdulmohsin, I., Parthasarathy, N., Evans, T., Beyer, L., Xia, Y., Mustafa, B., Hénaff, O., Harmsen, J., Steiner, A., Zhai, X.: Siglip 2: Multilingual vision-language encoders with improved semantic understanding, localization, and dense features (2025), \url{https://arxiv.org/abs/2502.14786}

\bibitem{wang2024mr}
Wang, G., Wei, X., Liu, J., Zhang, R., Zhang, Y., Zhang, K., Chong, M., Zhang, S.: {MR-MLLM}: Mutual reinforcement of multimodal comprehension and vision perception (2024), \url{https://arxiv.org/abs/2406.15768}

\bibitem{wang2026locateanything}
Wang, S., Liu, S., Kuang, Y., Wei, X., Liu, Y., Li, Z., Man, Y., Chen, G., Tao, A., Liu, G., Kautz, J., Zhang, L., Yu, Z.: Locateanything: Fast and high-quality vision-language grounding with parallel box decoding (2026), \url{https://arxiv.org/abs/2605.27365}

\bibitem{wang2025unifiedmultimodalchainofthoughtreward}
Wang, Y., Li, Z., Zang, Y., Wang, C., Lu, Q., Jin, C., Wang, J.: Unified multimodal chain-of-thought reward model through reinforcement fine-tuning (2025), \url{https://arxiv.org/abs/2505.03318}

\bibitem{wang2026unifiedrewardmodelmultimodal}
Wang, Y., Zang, Y., Li, H., Jin, C., Wang, J.: Unified reward model for multimodal understanding and generation (2026), \url{https://arxiv.org/abs/2503.05236}

\bibitem{wang2025videoverse}
Wang, Z., Wei, X., Li, B., Guo, Z., Zhang, J., Wei, H., Wang, K., Zhang, L.: {VideoVerse}: Does your {T2V} generator have world model capability to synthesize videos? (2025), \url{https://arxiv.org/abs/2510.08398}

\bibitem{wang2026timecausality}
Wang, Z., Zhang, S., Tang, C., Wang, K.: {TimeCausality}: Evaluating the causal ability in time dimension for vision language models (2025), \url{https://arxiv.org/abs/2505.15435}

\bibitem{wei2026mico}
Wei, X., Cen, K., Wei, H., Guo, Z., Li, B., Wang, Z., Zhang, J., Zhang, L.: {MICo-150K}: A comprehensive dataset advancing multi-image composition. In: Proceedings of the IEEE/CVF Conference on Computer Vision and Pattern Recognition (CVPR). pp. 29695--29706 (June 2026)

\bibitem{wiles2025revisitingtexttoimageevaluationgecko}
Wiles, O., Zhang, C., Albuquerque, I., Kajić, I., Wang, S., Bugliarello, E., Onoe, Y., Papalampidi, P., Ktena, I., Knutsen, C., Rashtchian, C., Nawalgaria, A., Pont-Tuset, J., Nematzadeh, A.: Revisiting text-to-image evaluation with gecko: On metrics, prompts, and human ratings (2025), \url{https://arxiv.org/abs/2404.16820}

\bibitem{wu2025qwenimagetechnicalreport}
Wu, C., Li, J., Zhou, J., Lin, J., Gao, K., Yan, K., ming Yin, S., Bai, S., Xu, X., Chen, Y., Chen, Y., Tang, Z., Zhang, Z., Wang, Z., Yang, A., Yu, B., Cheng, C., Liu, D., Li, D., Zhang, H., Meng, H., Wei, H., Ni, J., Chen, K., Cao, K., Peng, L., Qu, L., Wu, M., Wang, P., Yu, S., Wen, T., Feng, W., Xu, X., Wang, Y., Zhang, Y., Zhu, Y., Wu, Y., Cai, Y., Liu, Z.: Qwen-image technical report (2025), \url{https://arxiv.org/abs/2508.02324}

\bibitem{wuJanusDecouplingVisual2024}
Wu, C., Chen, X., Wu, Z., Ma, Y., Liu, X., Pan, Z., Liu, W., Xie, Z., Yu, X., Ruan, C., Luo, P.: Janus: {{Decoupling Visual Encoding}} for {{Unified Multimodal Understanding}} and {{Generation}} (Oct 2024). \doi{10.48550/arXiv.2410.13848}

\bibitem{wu2023humanpreferencescorev2}
Wu, X., Hao, Y., Sun, K., Chen, Y., Zhu, F., Zhao, R., Li, H.: Human preference score v2: A solid benchmark for evaluating human preferences of text-to-image synthesis (2023), \url{https://arxiv.org/abs/2306.09341}

\bibitem{wu2024conceptmixcompositionalimagegeneration}
Wu, X., Yu, D., Huang, Y., Russakovsky, O., Arora, S.: Conceptmix: A compositional image generation benchmark with controllable difficulty (2024), \url{https://arxiv.org/abs/2408.14339}

\bibitem{xieSANA15Efficient2025}
Xie, E., Chen, J., Zhao, Y., Yu, J., Zhu, L., Wu, C., Lin, Y., Zhang, Z., Li, M., Chen, J., Cai, H., Liu, B., Zhou, D., Han, S.: {{SANA}} 1.5: {{Efficient Scaling}} of {{Training-Time}} and {{Inference-Time Compute}} in {{Linear Diffusion Transformer}} (Mar 2025). \doi{10.48550/arXiv.2501.18427}

\bibitem{xieShowoOneSingle2024}
Xie, J., Mao, W., Bai, Z., Zhang, D.J., Wang, W., Lin, K.Q., Gu, Y., Chen, Z., Yang, Z., Shou, M.Z.: Show-o: {{One Single Transformer}} to {{Unify Multimodal Understanding}} and {{Generation}} (Oct 2024). \doi{10.48550/arXiv.2408.12528}

\bibitem{xie2025showo2improvednativeunified}
Xie, J., Yang, Z., Shou, M.Z.: Show-o2: Improved native unified multimodal models (2025), \url{https://arxiv.org/abs/2506.15564}

\bibitem{xin2025luminamgpt20standaloneautoregressive}
Xin, Y., Yan, J., Qin, Q., Li, Z., Liu, D., Li, S., Huang, V.S.J., Zhou, Y., Zhang, R., Zhuo, L., Han, T., Sun, X., Luo, S., Wang, M., Fu, B., Cao, Y., Li, H., Zhai, G., Liu, X., Qiao, Y., Gao, P.: Lumina-mgpt 2.0: Stand-alone autoregressive image modeling (2025), \url{https://arxiv.org/abs/2507.17801}

\bibitem{xu2025visionrewardfinegrainedmultidimensionalhuman}
Xu, J., Huang, Y., Cheng, J., Yang, Y., Xu, J., Wang, Y., Duan, W., Yang, S., Jin, Q., Li, S., Teng, J., Yang, Z., Zheng, W., Liu, X., Ding, M., Zhang, X., Gu, X., Huang, S., Huang, M., Tang, J., Dong, Y.: Visionreward: Fine-grained multi-dimensional human preference learning for image and video generation (2025), \url{https://arxiv.org/abs/2412.21059}

\bibitem{yang2025longt2ibenchbenchmarkevaluatinglong}
Yang, Z., Gu, T., Wang, J., Lin, F., Sheng, X., Chen, P., Li, L.: Longt2ibench: A benchmark for evaluating long text-to-image generation with graph-structured annotations (2025), \url{https://arxiv.org/abs/2512.09271}

\bibitem{ye2025echo4oharnessingpowergpt4o}
Ye, J., Jiang, D., Wang, Z., Zhu, L., Hu, Z., Huang, Z., He, J., Yan, Z., Yu, J., Li, H., He, C., Li, W.: Echo-4o: Harnessing the power of gpt-4o synthetic images for improved image generation (2025), \url{https://arxiv.org/abs/2508.09987}

\bibitem{zhang2025mavis}
Zhang, R., Wei, X., Jiang, D., Guo, Z., Li, S., Zhang, Y., Tong, C., Liu, J., Zhou, A., Wei, B., Zhang, S., Gao, P., Li, C., Li, H.: {MAVIS}: Mathematical visual instruction tuning with an automatic data engine. In: International Conference on Learning Representations (2025)

\bibitem{zhang2025baserewardstrongbaselinemultimodal}
Zhang, Y.F., Yang, H., Zhang, H., Shi, Y., Chen, Z., Tian, H., Fu, C., Wang, H., Wu, K., Cui, B., Wang, X., Pan, J., Wang, H., Zhang, Z., Wang, L.: Basereward: A strong baseline for multimodal reward model (2025), \url{https://arxiv.org/abs/2509.16127}

\bibitem{zhao2025lexartrethinkingtextgeneration}
Zhao, S., Wu, Q., Li, X., Zhang, B., Li, M., Qin, Q., Liu, D., Zhang, K., Li, H., Qiao, Y., Gao, P., Fu, B., Li, Z.: Lex-art: Rethinking text generation via scalable high-quality data synthesis (2025), \url{https://arxiv.org/abs/2503.21749}

\bibitem{zhou2022simplemultidatasetdetection}
Zhou, X., Koltun, V., Krähenbühl, P.: Simple multi-dataset detection (2022), \url{https://arxiv.org/abs/2102.13086}

\bibitem{zhuoLuminaNextMakingLuminaT2X2024}
Zhuo, L., Du, R., Xiao, H., Li, Y., Liu, D., Huang, R., Liu, W., Zhao, L., Wang, F.Y., Ma, Z., Luo, X., Wang, Z., Zhang, K., Zhu, X., Liu, S., Yue, X., Liu, D., Ouyang, W., Liu, Z., Qiao, Y., Li, H., Gao, P.: Lumina-{{Next}}: {{Making Lumina-T2X Stronger}} and {{Faster}} with {{Next-DiT}} (Jun 2024). \doi{10.48550/arXiv.2406.18583}

\end{thebibliography}
